\documentclass[journal]{IEEEtran}
\usepackage{times}
\usepackage{epsfig}
\usepackage{amssymb}
\usepackage{multirow}
\usepackage{xspace}
\usepackage{xcolor}
\usepackage{soul}
\sethlcolor{green}

\newcommand{\revised}[1]{#1}
\usepackage[normalem]{ulem}

\makeatletter
\DeclareRobustCommand\onedot{\futurelet\@let@token\@onedot}
\def\@onedot{\ifx\@let@token.\else.\null\fi\xspace}

\def\eg{\emph{e.g}\onedot} 
\def\ie{\emph{i.e}\onedot} 
 
\def\etc{\emph{etc}\onedot} 
\def\wrt{w.r.t\onedot} 
\def\etal{\emph{et al}\onedot}
\makeatother

\ifCLASSINFOpdf
   \usepackage{graphicx}
   \usepackage{subfigure}
\else
\fi
\usepackage{amsmath}
\usepackage{url}
\begin{document}
\title{PDNet: Toward Better One-Stage Object Detection With Prediction Decoupling}

% \author{Michael~Shell,~\IEEEmembership{Member,~IEEE,}
%         John~Doe,~\IEEEmembership{Fellow,~OSA,}
%         and~Jane~Doe,~\IEEEmembership{Life~Fellow,~IEEE}% <-this % stops a space
% \thanks{M. Shell was with the Department
% of Electrical and Computer Engineering, Georgia Institute of Technology, Atlanta,
% GA, 30332 USA e-mail: (see http://www.michaelshell.org/contact.html).}% <-this % stops a space
% \thanks{J. Doe and J. Doe are with Anonymous University.}% <-this % stops a space
% \thanks{Manuscript received ; revised .}}

\author{Li~Yang,
        Yan~Xu,
        Shaoru~Wang,
        Chunfeng~Yuan,
        Ziqi~Zhang,
        Bing~Li,
        and~Weiming~Hu%<-this % stops a space
\thanks{Li Yang, Shaoru Wang, and Ziqi Zhang are with the National Laboratory of Pattern Recognition, Institute of Automation, Chinese Academy of Sciences, and also with the School of Artificial Intelligence, University of Chinese Academy of Sciences, Beijing, China (e-mail: li.yang@nlpr.ia.ac.cn; wangshaoru2018@ia.ac.cn; zhangziqi2017@ia.ac.cn)}
\thanks{Yan Xu is with the Department of Electronic Engineering, The Chinese University of Hong Kong, Hong Kong, China (e-mail: yanxu@link.cuhk.edu.hk)}
\thanks{Chunfeng Yuan is with the National Laboratory of Pattern Recognition, Institute of Automation, Chinese Academy of Sciences, Beijing 100190, China (e-mail: cfyuan@nlpr.ia.ac.cn) (Corresponding author)}
\thanks{Bing Li is with the National Laboratory of Pattern Recognition, Institute of Automation, Chinese Academy of Sciences, Beijing 100190, China, and also with PeopleAI, Inc. (e-mail: bli@nlpr.ia.ac.cn)}
\thanks{Weiming Hu is with the CAS Center for Excellence in Brain Science and Intelligence Technology, also with the National Laboratory of Pattern Recognition, Institute of Automation, Chinese Academy of Sciences, and also with the School of Artificial Intelligence, University of Chinese Academy of Sciences, Beijing 100190, China (e-mail: wmhu@nlpr.ia.ac.cn)}
}


% \markboth{This is the new manuscript with the revised parts in bold.}%
% \markboth{This is the new manuscript.}%
\markboth{}%
{Shell \MakeLowercase{\textit{et al.}}: Bare Demo of IEEEtran.cls for IEEE Journals}



\maketitle

\begin{abstract}

%\sout{Recent one-stage object detectors follow a per-pixel prediction strategy to infer the category scores and boundary locations at each spatial grid that possibly presents an object. }
Recent one-stage object detectors follow a per-pixel prediction approach that predicts both the object category scores and boundary positions from every single grid location. %which substantially relates each object with the 2D grids and predicts the object category and boundary locations all from a single grid location. 
%\sout{However, it may leads to sub-optimal results as the most suitable positions for predicting different targets, \ie the category or boundaries, are not necessarily the same.}
However, the most suitable positions for inferring different targets, \ie, the object category and boundaries, are generally different. 
Predicting all these targets from the same grid location thus may lead to sub-optimal results. 
%\sout{In this paper, we propose a prediction decoupling mechanism to separate the targets to different locations, and re-collect predictions for each of them from the dense classification or regression maps.}
% In this paper, we analyze the suitable inference positions for different targets. 
%In this paper, we analyze the suitability of inference positions for different targets and propose the PDNet with a prediction decoupling mechanism, which predicts different targets from separate locations, and collects the predictions to form the final detection results. %during inference.
%\sout{suitability of inference positions}
In this paper, we analyze the suitable inference positions for object category and boundaries, and propose a prediction-target-decoupled detector named PDNet to establish a more flexible detection paradigm. %\sout{address the limitations above}\sout{ to extend the above prediction manners}
Our PDNet with the prediction decoupling mechanism encodes different targets separately in different locations. % \sout{predicts}from different locations 
A learnable prediction collection module is devised with two sets of dynamic points, \ie, dynamic boundary points and semantic points, to collect and aggregate the predictions from the favorable regions for localization and classification. 
We adopt a two-step strategy to learn these dynamic point positions, where the prior positions are estimated for different targets first, and the network further predicts residual offsets to the positions with better perceptions of the object properties.
Extensive experiments on the MS COCO benchmark demonstrate the effectiveness and efficiency of our method. 
With a single \revised{ResNeXt-64x4d-101-DCN} as the backbone, our detector achieves \revised{50.1} AP with single-scale testing, which outperforms the state-of-the-art methods by an appreciable margin under the same experimental settings. 
Moreover, our detector is highly efficient as a one-stage framework.
Our code is public at \url{https://github.com/yangli18/PDNet}.
\end{abstract}


% \fi
\IEEEpeerreviewmaketitle

\section{Introduction}

%\IEEEPARstart{T}{his}

%1. (one-stage detectors; two-stage detectors).

%\sout{detection has been developing rapidly in recent years. } 
\IEEEPARstart{O}{bject} detection is a fundamental problem in computer vision aiming to localize and classify objects in digital images. 
In terms of the prediction stages needed by the detector, %\sout{The}
existing object detection networks %framework \sout{detection methods} 
can be generally categorized as the one-stage method~\cite{liu2016ssd,lin2017focal} and the two-stage method~\cite{ren2015faster,dai2016r}.
The one-stage detectors directly produce the classification and localization results from dense grid points in one shot without \revised{an} explicit feature alignment procedure, %\sout{for further refinement}, 
while the two-stage methods include an additional stage of RoI feature extraction to %\sout{achieve accurate detection} further 
improve the detection performance in a coarse-to-fine manner~\cite{ren2015faster}.

\begin{figure}[t]
	\centering
	\includegraphics[width=0.99\columnwidth, trim=72 12 57 18,clip]{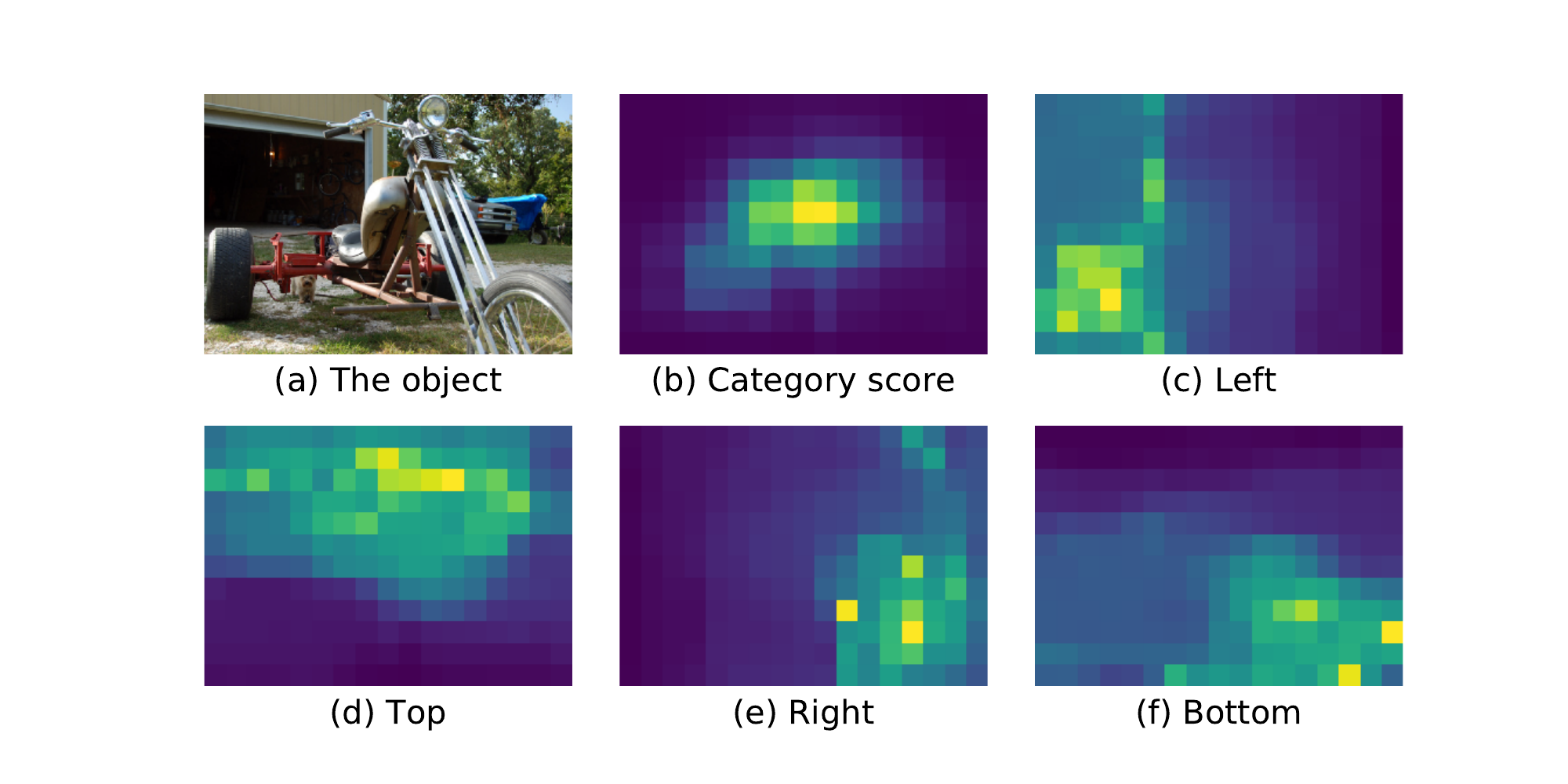}
	\caption{
The accuracy maps of per-pixel classification and localization results from the current one-stage detector~\cite{tian2019fcos}. 
\revised{The brighter areas produce more accurate predictions for the respective targets (including the object category and the left, top, right, bottom edges of the bounding box).} %\sout{the respective} target\sout{s}
\revised{The different accuracy distributions of these target predictions motivate us to decouple the localization and classification predictions.} 
}
\label{figure_fcos_pred}
\end{figure}

Due to the structural efficiency and competitive performance, one-stage methods have received great attention. 
The recent cutting-edge methods~\cite{tian2019fcos,kong2020foveabox,zhu2019feature,zhu2020soft} 
abandon the anchor box references and develop more straightforward detection frameworks that perform per-pixel classification and regression.  
For each output grid belonging to an object, these detectors predict the category scores and the current offsets %\sout{relative}
to the leftmost, topmost, rightmost, and bottommost sides of the object. %let each grid belonging to an object to predict
However, given the various object poses and shapes, it could be challenging for the features located at a single grid to accurately perceive the object category and four sides of the bounding box altogether. Intuitively, for example, it could be more difficult for the positions near the object boundary to perceive the semantic information than the positions inside. It could also be less accurate to regress the left border of the object from the positions near the right side. 
\revised{To validate our conjecture, we analyze the predictions of a conventional one-stage detection framework~\cite{tian2019fcos}. 
As shown in Fig.~\ref{figure_fcos_pred}, we visualize the accuracy of classification and localization  predictions at different locations around the target object. 
Concretely, for classification, we visualize the estimated confidence scores on the ground-truth category. For localization, we show the inverse regression errors of the bounding-box edges.} 
From the accuracy maps, we can tell that the best positions for predicting different targets vary as expected. 
For the object instance, the regions near each object boundary tend to produce more accurate localization results, while the positions close to the semantic area tend to have wise predictions for the object category. 
These observations naturally raise a question: is it possible to efficiently separate the prediction targets and obtain the predictions for each target at their respective favorable positions in one shot?

To this end,  we propose a \textbf{P}rediction-target-\textbf{D}ecoupled detection \textbf{Net}work (PDNet), where different targets are inferred separately at their corresponding proper positions. 
Specifically, unlike the previous one-stage methods that classify and localize an object instance from the same grid location in the feature map, we propose a prediction decoupling mechanism to separate the prediction targets as the object category and four sides of the object bounding box (in an offset manner), which are separately encoded at different locations by the network. 
% Then, 
To obtain the final detection results, we devise a learnable prediction collection module to collect and aggregate these intermediate predictions for different targets from different locations. 
Moreover, we analyze the suitable inference positions for localization and classification 
and propose two sets of dynamic points, \ie, dynamic boundary points and semantic points, %\sout{to model these locations respectively}
to pinpoint these \revised{positions} respectively. %\sout{locations}% 
To this end, we introduce a two-step dynamic point generation strategy to facilitate the learning of dynamic points. % \sout{learning strategy} 
The network first roughly estimates the prior positions for different targets to initialize the dynamic points, and then further predicts the residual offsets to shift the dynamic points towards the positions where the object boundaries or semantic properties can be better perceived.

With the established dynamic points, we can flexibly collect the localization and classification results from the respective appropriate locations, 
to better exploit the one-stage detector's potential. 
The proposed prediction decoupling mechanism naturally focuses different positions on the prediction of different targets, which empirically addresses the limitation of conventional one-stage detection approaches. % and thus achieve better performance. 
It is worth mentioning that the prediction decoupling and collection process is lightweight and helps keep the one-stage detection framework's efficiency advantage. %to keep the efficiency advantage of the one-stage detection framework.

%4. Contribution
To summarize, the contributions of this work are:
\begin{itemize}
\item We analyze the dense predictions of the conventional one-stage detector and find that the best positions for inferring the object category and boundary positions are different. %\sout{prediction results}
Inspired by the phenomena, we propose the PDNet with a prediction decoupling mechanism to flexibly collect and aggregate the predictions for different targets from different locations. 
\item We devise two sets of dynamic points, \ie, dynamic boundary points and semantic points, 
and propose a two-step dynamic point generation strategy to facilitate the learning of suitable point positions for localization and classification.
\item Without bells and whistles, our method achieves state-of-the-art performance on the MS COCO benchmark. With a single \revised{ResNeXt-64x4d-101-DCN} as the backbone, our detector achieves \revised{50.1} AP with single-scale testing, outperforming the other methods by an appreciable margin under the same experimental settings. 
\end{itemize}

%------------------------------------------------------------------------
\section{Related work}

In this section, we briefly introduce the two-stage and one-stage object detection methods, and also present various detection head designs for object localization and classification.

\subsection{Two-stage Detection} %\subsection{Two-stage Methods}

\revised{
Faster R-CNN~\cite{ren2015faster} establishes the foundation of the modern two-stage detection framework. 
It first uses a region proposal network (RPN) to generate object region proposals, which are then processed by a region-based convolutional neural network (R-CNN)~\cite{girshick2014rich,girshick2015fast} to perform classification and localization refinement. 
Based on this detection pipeline, many methods have been proposed for performance improvement from various aspects. 
\cite{dai2016r,lin2017feature,li2019scale,cai2018cascade} propose to improve the network design; \cite{he2017mask,dai2017deformable,zhu2019deformable} establish better object feature representations; \cite{wang2019region,vu2019cascade} develop better region proposals; \cite{shrivastava2016training,singh2018analysis,jiang2018acquisition,pang2019libra,zhang2020dynamic,he2019bounding} design better training strategies and loss functions. While this two-stage detection paradigm has performance advantages, it increases the complexity of the network structure, which motivates the development of efficient one-stage detectors. 
}

\subsection{One-stage Detection} %\subsection{One-stage Methods}

The \revised{seminal works} SSD~\cite{liu2016ssd} and RetinaNet~\cite{lin2017focal} establish simple and effective one-stage detection frameworks. %\sout{emergence of}\yx{effective but still powerful} 
These methods preset anchor boxes of various sizes at each grid location.
\revised{During inference, they directly refine the anchor boxes' locations and classify them as objects or backgrounds.} 
\revised{Following these pioneering frameworks, many great works have been proposed with significant improvements~\cite{fu2017dssd,zhang2018single,zhang2019freeanchor,ke2020multiple,zhang2020bridging,li2020generalized}. 
Besides the above anchor-based one-stage methods, another branch of one-stage detectors gets rid of the anchors~\cite{redmon2016you,huang2015densebox}. 
Recently, the anchor-free methods~\cite{tian2019fcos,kong2020foveabox,zhu2019feature,zhu2020soft} have achieved comparable or even better performance compared with the anchor-based methods.} 
Without \revised{the} anchor boxes, 
they directly predict the category scores and the offsets to four sides of the object bounding box at each \revised{feature map grid.} 
However, they infer all the properties (the location and the category) of an object from the same grid location, which may \revised{result in compromised results of localization and classification.} 
In our work, we propose to more flexibly collect the localization and classification predictions from different positions \revised{to alleviate this dilemma.}

\subsection{Detection Head Design for Localization and Classification}

\revised{The detection head is the key component for accurate object detection. Various detection heads have been proposed to push the performance boundary.} % \sout{of object detection}. 
Based on the two-stage detection pipeline, Double-Head~\cite{wu2020rethinking} \revised{devises different detection heads for classification and localization respectively.} 
TSD~\cite{song2020revisiting} \revised{proposes to generate} different region proposals for classification and localization. 
Grid R-CNN~\cite{lu2019grid} \revised{adopts a grid guided localization mechanism for accurate detection.} 
SABL~\cite{wang2020side} constructs side-aware features to localize each boundary in a coarse-to-fine manner. 
Unlike these methods that rely on RoI features for prediction, we inherit the scheme of one-stage methods and collect the localization and classification \revised{predictions} at \revised{more appropriate} positions to achieve accurate and efficient detection.

\revised{Some recent works incorporate object feature extraction into one-stage detectors for accurate localization and classification.} 
RepPoints~\cite{yang2019reppoints,chen2020reppoints} formulates the object as a set of representative points for feature sampling. 
AlignDet~\cite{chen2019revisiting} proposes RoIConv to align the convolution features with \revised{the} object proposals \revised{for detection.} %\sout{to make predictions.} 
Based on the \revised{detection results} of FCOS~\cite{tian2019fcos}, BorderDet~\cite{qiu2020borderdet} gathers the border features to refine the classification and localization predictions. %the detections of FCOS.
While exhibiting better performance, these methods all need additional detection branches to make predictions on \revised{the} extracted features, which may sacrifice inference efficiency. %, which tends to sacrifice the efficiency and complicate the model pipeline.
\revised{In contrast}, we only collect predictions from the regression and classification maps of the conventional one-stage detection pipeline, which achieves higher accuracy while keeping the efficiency advantage. 

\revised{Another family of object detection methods follows a bottom-up approach to localize and classify objects.} CornerNet~\cite{law2018cornernet} proposes to detect an object bounding box as a pair of keypoints, \revised{\ie,} the top-left corner and the bottom-right corner. It first predicts the heatmaps of corner points for different categories and then groups the corner points with similar embeddings to form the detection results. CenterNet~\cite{duan2019centernet} extends CornerNet by introducing the detection of center keypoints to improve accuracy and recall. Zhou~\etal~\cite{zhou2019objects} \revised{propose to directly detect} the object centers and \revised{regress} the object sizes. ExtremeNet~\cite{zhou2019bottom} detects four extreme points and one center point of objects and proposes center grouping to produce the detection results. 
Compared with these methods, we \revised{generate} dynamic points to collect localization and classification \revised{predictions}, and no additional embedding or grouping operations are required during post-processing.

%------------------------------------------------------------------------
\section{Our method}
In this section, we first analyze the suitability of inference positions for different targets, \ie, object boundaries and categories, in one-stage object detection. 
Based on the analysis, we propose a prediction decoupling mechanism to focus different locations on the prediction of different targets. 
Then, we further devise dynamic points to locate the appropriate positions for collecting predictions. Finally, we elaborate on the network architecture and the details of training and inference. %\sout{model} \sout{predicting}

\subsection{Analysis} \label{Analysis}

The conventional state-of-the-art one-stage detectors generally infer the object locations and categories from the central areas of objects~\cite{tian2019fcos,kong2020foveabox}. 
However, as shown in Fig.~\ref{figure_fcos_pred}, the most suitable positions for inferring different targets might differ as well. 
Directly inferring the object category and boundaries from the same grid location, which has been widely adopted by the conventional one-stage methods, might require rethinking. 
To find the optimal inference locations for different targets and build up a more powerful detector, we first conduct some experiments to evaluate the performance of dense classification and regression predictions in object regions.

\begin{figure}[t]
    \setlength{\abovecaptionskip}{1pt}
	\centering
	\includegraphics[width=0.99\columnwidth, trim=10 0 10 0,clip]{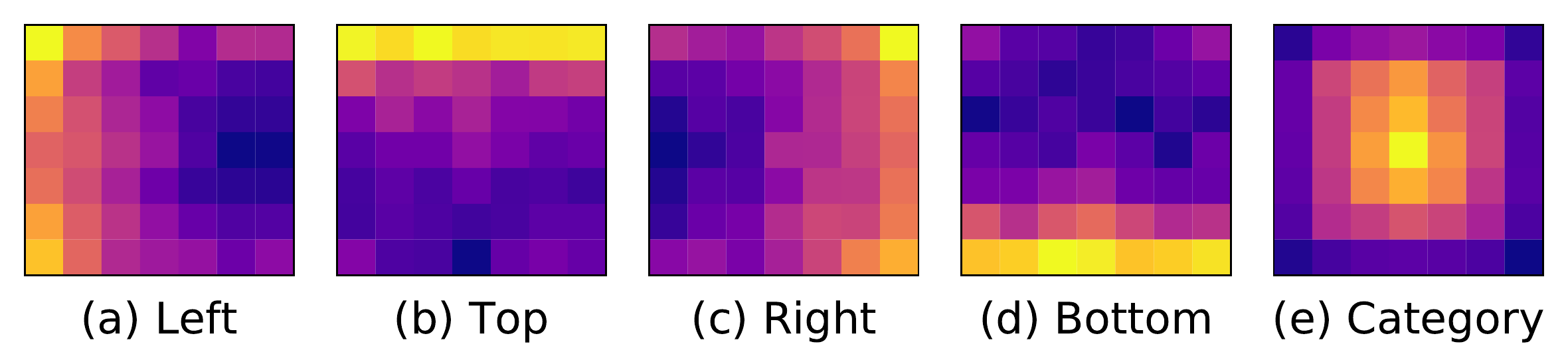} 
	% Reduce the figure size so that it is slightly narrower than the column.
	\caption{\revised{
	The spatial distributions of the locations that produce the best estimates for the respective targets (visualized as 2D histograms). 
	 We analyze the best estimates for different targets separately (including the left, top, right, bottom bounding box edges and the object category). 
	 The brighter region implies a higher probability of producing accurate estimates. 
    This analysis is conducted on the MS COCO validation set~\cite{lin2014microsoft}.
}
	}
	\label{fig_pred_distribution}
% 	\vspace{-0.28cm}
\end{figure}

We train the one-stage detection network~\cite{tian2019fcos}, where we assign all the grids inside the object bounding box as positive samples for object localization and classification during training. 
Then, we evaluate the detection network on the validation set~\cite{lin2014microsoft} and analyze its dense detection results. 
Specifically, for each object instance with ground-truth bounding box annotation, we sample the grids containing detection results that have IoU $>\!0.5$ \wrt the corresponding ground-truth bounding box. \revised{Then, we statistically analyze the grid locations where the %\sout{most accurate predictions}
best estimates are generated for different targets (\ie, the grids producing the highest confidence scores on the ground-truth category or the most precise localization results for each side).} %\sout{\ie, the object bounding box boundaries and the object category}
\revised{The 2D histograms in Fig.~\ref{fig_pred_distribution} exhibit the spatial distributions of these favorable grid locations (normalized by the object bounding box size) for different targets}.
It can be observed that the regions around the object are more suitable for inferring the locations of four sides of the object bounding box, while the areas covering the object tend to have higher confidence in category classification. 
This phenomenon reveals the different key factors for localization and classification, which also meets the intuition that it would be easier for the areas near the object contour to perceive the boundary, while the object category needs to be identified from the inner semantic regions of the object. 
Therefore, we argue that the predictions for different targets should be obtained from their more appropriate locations. 
In the following sections, we will discuss the prediction decoupling mechanism and propose a unified detection framework with the prediction decoupling to push the one-stage detection performance boundary.

\begin{figure*}[t]
	\centering
	\includegraphics[width=0.988\textwidth]{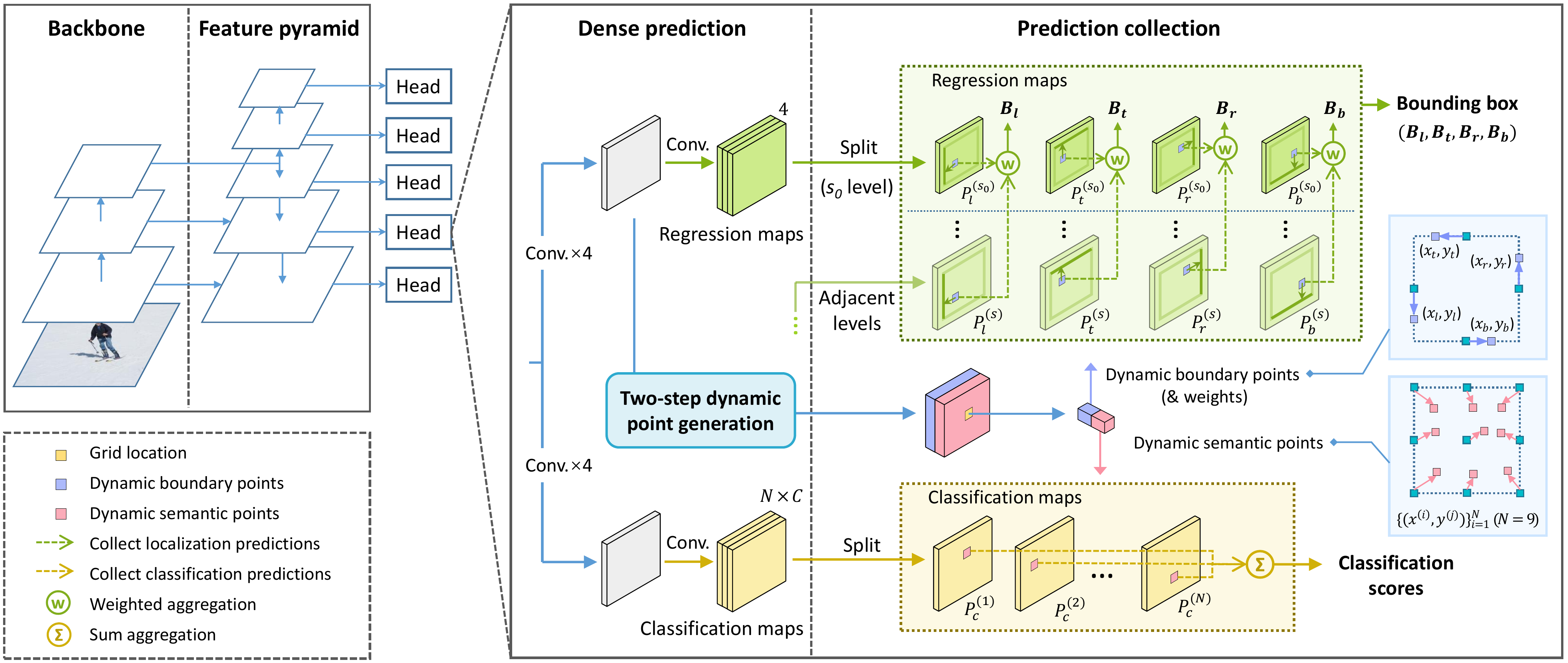}
	% Reduce the figure size so that it is slightly narrower than the column.
	%\vspace{0.2cm}
	\caption{
	The overall network architecture of PDNet. 
	Based on the feature extraction backbone and the feature pyramid network (FPN), the PDNet extends multiple detection heads from the FPN for multi-scale dense detection. %dense prediction. 
	In the detection head, the dense prediction maps for classification and localization are first produced in the \textbf{dense prediction} step, similar to most of the conventional one-stage methods. 
	%\sout{The difference is that }
	Our prediction maps are split along the channel dimension, where different channels encode the corresponding different targets for each location. 
	Concretely, the regression map slices $\{P_\tau^{(s)}\}_{s\in \mathbb{N}(s_0)}$ (in green), where $\tau\in \{l,t,r,b\}$, contain the dense predictions of the relative offsets to the four sides of object bounding boxes, %\sout{$\mathbf{B}_l, \mathbf{B}_t, B_r$ and $B_b$}, 
	while the classification map slices $\{P_c^{(i)}\}_{i=1}^N$ (in yellow) hold the dense classification scores %\sout{corresponding to}
	of different semantic regions. %dynamic semantic points. 
    After obtaining these dense predictions, for each grid location, we perform \textbf{prediction collection} guided by the two sets of dynamic points (from the \textbf{{two-step dynamic point generation module}}) %\sout{for each grid location,} %(generated from the two-step module)
    to obtain the classification scores and the bounding box $(B_l, B_t, B_r, B_b)$ by gathering predictions from the respective favorable positions.
	}
	\label{fig_network}
\end{figure*}

\subsection{Prediction Decoupling} \label{Prediciton Decoupling}

Given an input image, one-stage detection networks~\cite{tian2019fcos,kong2020foveabox} generate multi-level dense prediction maps containing the category scores and the boundary locations of objects.
We let $P_\tau$ denote a prediction map for a specific target $\tau \in\{c,l,t,r,b\}$, \ie, either the category $c$ or the boundary locations for each side indexed by $l, t, r,$ and $b$.
In the conventional one-stage paradigm, for the object corresponding to the grid $(x, y)$, %{for the object to be detected at the grid location $(x, y)$},%for a grid located at $(x, y)$ corresponding to an object, %that likely corresponds to an object, 
the prediction result $R_{\tau}(x, y)$ for each target $\tau$ %different target $\tau \in\{c,l,t,r,b\}$ 
identically comes from the same location in the prediction map $P_{\tau}$, \ie, $R_{\tau}(x, y)=P_{\tau}(x, y)$.
However, as mentioned above, such prediction manners %collection 
tend to be sub-optimal, since the prediction results for different targets may need to be obtained from different locations. 
To this end, we need to learn a map $G_\tau$ for each target $\tau$, with which we can %are able to 
locate the suitable location $(x', y')$ to collect the predictions for the object at the current location $(x, y)$. %\sout{for different targets}.
We formulate this collection process as:
\begin{equation}\label{eq:simple_decouple}
\left\{
\begin{aligned}
 	(x', y') &= G_\tau(x, y) \\
 	R_\tau(x, y) &= P_\tau(x', y')
\end{aligned}
\right.,
\end{equation}
where the prediction result for the object at $(x, y)$ can be collected flexibly from the more appropriate location $(x', y')$ on the prediction map. %\sout{among}, \sout{achieving prediction decoupling for different targets}. 
The above operations essentially assign the tasks of different target predictions to the respective more advantageous locations, and we thus named this mechanism \textit{prediction decoupling}.

Eq.~\eqref{eq:simple_decouple} only allows flexible prediction collection for different targets. 
However, for each target, %\sout{more than one collected prediction} 
multiple prediction results may need to be incorporated for better modeling. %to be better modeled.
Specifically, the object category may need the predictions from different semantic parts to jointly determine, while the boundary locations could be better estimated by choosing the predictions at %\sout{in the multi-scale predictions}
proper scale levels (from multi-scale localization predictions). 
Thus, we further extend the above formulation Eq.~\eqref{eq:simple_decouple} into a more general version that can utilize multiple predictions for each target: 
\begin{equation}
\left\{
\begin{aligned}
&\mathbf{X}' = G_\tau(x, y), \quad \mathbf{X}' \in \mathbb{R}^{K\times 2} \\
&R_\tau(x,y) =\Phi\big(\mathbf{P}_{\tau, \mathbf{X}'} \big)
\end{aligned}
\right.%.,
\end{equation}
Compared with Eq.~\eqref{eq:simple_decouple}, here we model each target by $K$ collected predictions from different locations, arranged in a matrix as $\mathbf{P}_{\tau, \mathbf{X}'}\!=\!$
\begin{footnotesize}$\left[P_\tau^{(1)}(x^{(1)},y^{(1)}), ..., P_\tau^{(K)}(x^{(K)},y^{(K)})\right]^{\mathrm{T}}$\end{footnotesize}
$\in \mathbb{R}^{K\times C}$, \revised{where $C$ denotes the dimension of the collected predictions.} 
Specifically, for each target $\tau \in\{c,l,t,r,b\}$, 
we generate the multiple collection locations as %get the collection locations as
$\mathbf{X}'=$
\begin{footnotesize}
$\left[(x^{(1)},y^{(1)}), ..., (x^{(K)},y^{(K)})\right]^{\mathrm{T}}$
\end{footnotesize}
$\in \mathbb{R}^{K\times 2}$
to obtain the predictions $\mathbf{P}_{\tau,\mathbf{X}'}$ from the respective prediction maps $\{P_\tau^{(i)}\}_{i=1}^K$ (for multiple levels or semantic parts, detailed in Section~\ref{Dynamic_points}). %, and then use 
Then, we use the aggregate function $\Phi(\cdot)$ to produce the final results. 
Note that for localization and classification targets, we propose different locations %$\{G_\tau^{(i)}\}_{i=1}^K$ 
to collect and aggregate predictions, which will be elaborated in Section~\ref{Dynamic_points}. 
The prediction collection process incurs almost negligible overhead, which can be easily integrated into the dense detection pipelines~\cite{tian2019fcos,kong2020foveabox}.

\subsection{Dynamic Points in Prediction Decoupling} \label{Dynamic_points}

We propose to establish two sets of dynamic points to locate the appropriate positions for predicting localization and classification targets. 
However, directly having the network learn such locations automatically may be difficult, and the optimization process can easily fall into local optimums.
To alleviate this, we propose a two-step dynamic point generation module %a two-step learning strategy 
that initializes the dynamic points to the prior positions for different targets and further shifts the points with the residual positional offsets predicted by the network. 
We will elaborate on the different dynamic point configurations for localization and classification separately in the ensuing parts.

\begin{figure*}[t]
\centering
\includegraphics[width=0.968\textwidth]{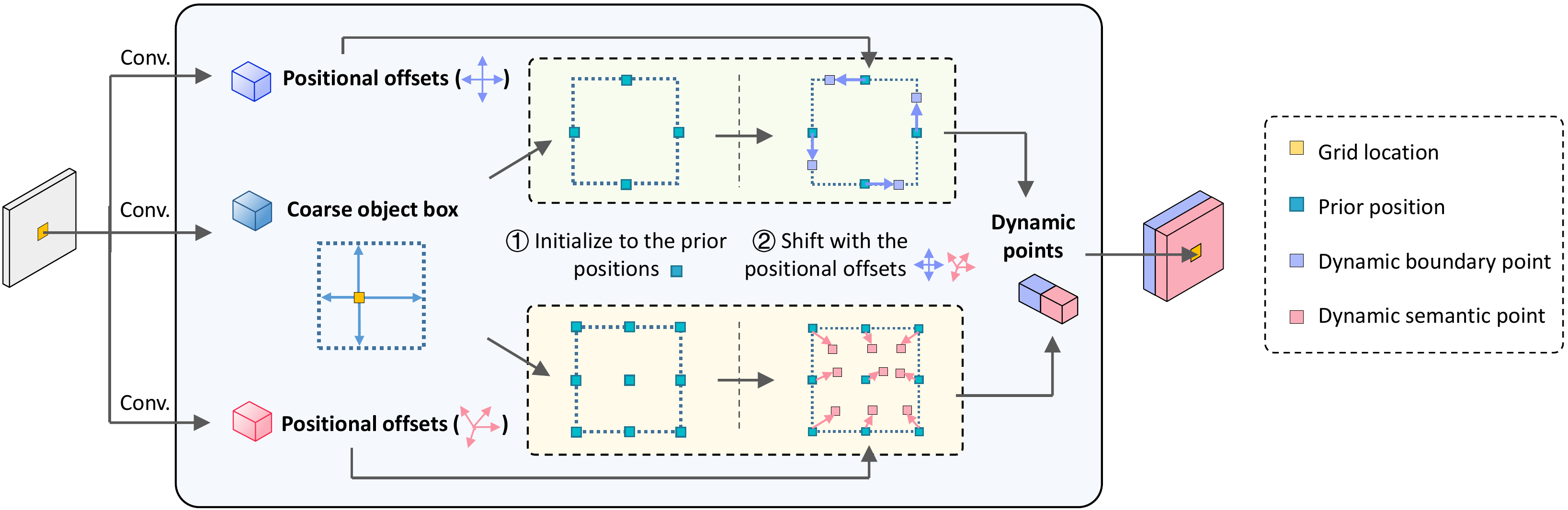}
% Reduce the figure size so that it is slightly narrower than the column.
\caption{
\revised{The two-step dynamic point generation module. For each grid location, the dynamic points are generated in a two-step manner: 1) We first estimate a coarse object box by a convolution layer, and initialize the dynamic points to some prior positions (\ie, the midpoints of the coarse box edges or the uniformly-distributed points in the coarse box). 
2) We estimate two sets of positional offsets with the convolution layers 
% (in parallel to the coarse box estimation ) 
to further shift these points towards better locations. 
% Specifically, 
In our pipeline, the dynamic boundary points (for localization) are shifted along the coarse box edges to better perceive object boundaries, while the dynamic semantic points (for classification) are shifted more flexibly to semantically representative regions.} %,focus on the object's
}
\label{fig_dyn_pt_module}
% \vspace{-0.28cm}
\end{figure*}

\subsubsection{Dynamic Boundary Points for Localization}\label{sec:localization}

\revised{As analyzed in Section~\ref{Analysis}, the areas near the object edges tend to be more suitable for object localization. Thus, we set several dynamic points in these areas to pinpoint the object bounding box. We refer to these points as dynamic boundary points in the following.} 
To effectively find the appropriate dynamic boundary points for each object, we decompose the point generation into two steps with the dynamic point generation module, which is branched from the network's dense prediction head \revised{(as shown in Fig.~\ref{fig_network}).} 
\revised{Specifically, at each grid location, 
we first estimate a coarse object box by predicting the offsets from the current grid to bounding box edges with a convolution layer. 
We initialize the dynamic boundary points at the midpoints of these coarse box boundaries, which are roughly close to the object edges 
(see the analyses in Section~\ref{experiment_loc_pt}). Thereafter, we further adjust the point locations along the coarse boundaries with the positional offsets generated by another convolution layer (parallel to the one estimating the coarse box). %previous
In this manner, the dynamic boundary points can be further pushed closer to the object edges and obtain the final position $(x_\tau, y_\tau)$ ($\tau\in\{l,t,r,b\}$). The suffixes here index the final dynamic point positions for the \textit{\textbf{l}eft}, \textit{\textbf{t}op}, \textit{\textbf{r}ight}, and \textit{\textbf{b}ottom} sides.
We demonstrate this two-step generation process of dynamic boundary points with the upper branch in Fig.~\ref{fig_dyn_pt_module} for better understanding.} %Fig.~\ref{fig_network} demonstrates an example for better understanding.

After having these optimized dynamic boundary points, we collect the respective regression predictions used to pinpoint the boundaries of the object bounding box. 
Let $P_{l}, P_{t}, P_{r}, P_{b}$ denote the respective dense regression maps \revised{containing} the per-pixel positional offsets \wrt the four box boundaries of objects. %(from each current point to the four box boundaries of objects). 
\revised{We collect the regression offsets from the dynamic boundary point locations $\{(x_\tau,y_\tau)\}$ ($\tau\!\in\!\{l,t,r,b\}$), obtaining $P_{l}(x_l, y_l)$, $P_{t}(x_t, y_t)$, $P_{r}(x_r, y_r)$, and $P_{b}(x_b, y_b)$, respectively. Thereafter, we add these collected offsets to the respective dynamic point locations to obtain the four boundaries of the bounding box $B$ as:}
% \sout{Then, the four boundaries of the final bounding box $B$ can be obtained as: }
\begin{equation}\label{eq_regression}
\begin{aligned}
B_l &= P_{l}(x_l, y_l) + x_l,\quad B_t = P_{t}(x_t, y_t)+ y_t, \\
B_r &= P_{r}(x_r, y_r) + x_r,\quad B_b = P_{b}(x_b, y_b) + y_b,
\end{aligned}
\end{equation}

Considering the various scales and aspect ratios of objects, the regression map of a specific scale level may be insufficient %{not be enough} %be hard
to perceive and localize the object boundaries well. We hence propose to choose the predictions at the regression maps of more suitable levels to collect the localization results. 
Concretely, when localizing the object corresponding to the level $s_0$, %FPN feature map of scale level $s_0$, 
we collect the predictions from the current scale level $s_0$ as well as the adjacent levels (denoted by $\mathbb{N}(s_0)$), and select the most confident localization predictions with a differentiable weighting mechanism. 
%{$\mathbf{P_\tau}\in \mathbb{R}^{K\times 1}$} \sout{arranged in a matrix, } \sout{each row}
With the collected predictions $\mathbf{P_\tau}\in \mathbb{R}^{K\times 1}$ (where each element is the prediction 
\begin{small}$ P_\tau^{(s)}(x_\tau^{(s)},y_\tau^{(s)}) $\end{small} collected from the level $s\!\in\!\mathbb{N}(s_0)$)
and the learned soft weights $\mathbf{W}_\tau \in \mathbb{R}^{K\times 1}$ for $K$ different levels ($K=|\mathbb{N}(s_0)|$), we take an aggregate function  
\begin{small}
$\Phi(\mathbf{P}_\tau;\mathbf{W}_\tau)=\textstyle \mathbf{W}^\mathrm{T}_\tau \mathbf{P}_\tau=\sum_{s\in \mathbb{N}(s_0)} W^{(s)}_\tau \cdot P^{(s)}_\tau(x_\tau^{(s)}, y_\tau^{(s)})$
\end{small}
to weight these collected offset predictions from multiple scale levels $\mathbb{N}(s_0)$, which essentially selects the suitable levels to obtain the final localization results, as shown in Fig.~\ref{fig_network}. 
By enhancing Eq.~\eqref{eq_regression} with this weighted multi-level aggregation, we have the regression equation for each side of the bounding box. For instance, the left boundary location is calculated as:
\begin{equation} \label{eq:dynamic_bbox}
\begin{small}
% B_l = \sum\nolimits_{s\in \mathbb{N}(s_0)}P_l^{(s)}(x_l^{(s)}, y_l^{(s)}) \cdot w_l^{(s)} + x_l
B_l = \Phi(\mathbf{P}_l;\mathbf{W}_l) + x_l=\!\sum\limits_{s\in \mathbb{N}(s_0)} \!W_l^{(s)}\cdot P_l^{(s)}(x_l^{(s)}, y_l^{(s)})  + x_l,
\end{small}
\end{equation}
%\end{small}
where the generated soft weights $W_l^{(s)}$ are normalized to satisfy $\sum_{s}W_l^{(s)}\!=\!1$. 
Compared with Eq.~\eqref{eq_regression}, for each obtained dynamic boundary point $(x_l^{(s_0)}, y_l^{(s_0)})$, we map it to the adjacent scale levels as $(x_l^{(s)}, y_l^{(s)})$ according to the interpolation rules, to fetch the corresponding prediction results.
During training, the dynamic boundary points are pushed towards the positions with better localization results by the regression loss $L_{reg}$ (Section~\ref{seq:training}). %\sout{applied on the final bounding boxes}
This enables the dynamic boundary points to flexibly adapt to %\yx{closely attached to} 
the object silhouettes, %\sout{edges}
which is crucial for achieving accurate bounding box prediction.

\subsubsection{Dynamic Semantic Points for Classification}\label{sec:classification}
From the analysis in Section~\ref{Analysis}, the inner regions of an object are more suitable for inferring the class labels, since these regions may contain richer semantic information helpful for determining the category. 
To efficiently pinpoint these semantic region positions, we define another set of dynamic points, named dynamic semantic points.   
\revised{As shown in the lower branch of Fig.~\ref{fig_dyn_pt_module}, we design the dynamic point generation module to generate these semantic points in a two-step manner similar to the dynamic boundary point generation. First, we take the previously estimated coarse object box (mentioned in Section~\ref{sec:localization}) as a reference and uniformly %\sout{borrow the coarse object box estimated in Section~\ref{sec:localization} and initially}
distribute $N$ points in the coarse box as the prior positions for the semantic points ($N=9$ for demonstration). These prior positions are expected to roughly cover different parts of the target object. %\sout{as shown in Fig.~\ref{fig_network} (right side, $N=9$ for demonstration)}
Then, in the dynamic point generation module, we estimate $N$ positional offsets to shift these points to positions that better perceive the semantic regions of the object, finally obtaining $\{(x^{(i)}, y^{(i)})\}_{i=1}^N$.} %\sout{with a convolution layer}

In our implementation, we predict $N$ classification maps $\{P_{c}^{(i)}\}_{i=1}^{N}$ (each with $C$ channels for $C$ classes) %(with $C$ channels each when $C$ classes exist) 
in parallel to model $N$ different semantic parts of objects. % separately. 
Each dynamic semantic point that represents a certain semantic part, is associated with a specific dense classification map. %Each dense classification map is associated to a specific dynamic semantic point. 
\revised{Specifically, after obtaining the position $(x^{(i)}, y^{(i)})$ of the $i$-th semantic point, we will collect the $C$-class scores voted by this point from its associated classification map $P_{c}^{(i)}$. The collected score vector is represented as $P_{c}^{(i)}(x^{(i)}, y^{(i)}) \in \mathbb{R}^{C}$. 
To jointly identify the object category from multiple semantic parts, we gather the voting score vectors from $N$ different point locations $\{(x^{(i)}, y^{(i)})\}_{i=1}^N$ as 
$\mathbf{P}_c=$
\begin{footnotesize}
$\left[P_c^{(1)}(x^{(1)},y^{(1)}), ..., P_c^{(N)}(x^{(N)},y^{(N)})\right]^{\mathrm{T}}$ \end{footnotesize}
$\in \mathbb{R}^{N\times C}$
and aggregate them by function 
\begin{small}
$\Phi(\mathbf{P}_c)= \mathbf{1}^\mathrm{T} \mathbf{P}_c= \sum_{i=1}^{N}P_{c}^{(i)}(x^{(i)}, y^{(i)})$ 
\end{small}
with a sigmoid function to produce the final $C$-class probability scores $s_c\in \mathbb{R}^C$}: %\sout{classification results}
\begin{equation}
%\begin{aligned}
\mathbf{s_c} = \frac{1}{1+\exp \big(-\sum_{i=1}^{N}P_{c}^{(i)}(x^{(i)}, y^{(i)}) \big)} \\
\end{equation}

Since the final \revised{category scores are} directly voted from the classification scores of each \revised{semantic} point, the classification loss can automatically drive the points towards the representative areas 
where the corresponding object category can be better perceived. 
This in turn helps us to collect better classification results from the optimized semantic points, making a more confident detection.

\subsection{The Network Architecture}
The overall architecture of our detection network is illustrated in Fig.~\ref{fig_network}. We employ a %\sout{similar structure}
paradigm similar to other one-stage detectors~\cite{lin2017focal,tian2019fcos}, including an image processing backbone~\cite{he2016deep}, a feature pyramid network~\cite{lin2017feature}, and multiple detection heads for multi-scale object detection. 
In each detection head, following the dense prediction convention, the %\sout{multi-convolution}
regression and classification branches produce dense prediction maps. 
The regression predictions (illustrated as green blocks) are divided along the channel dimension into four regression maps that contain the relative offsets to %\sout{localization of (\ie, the offsets to)}
four sides of objects respectively, which are used for locating the object bounding boxes.
Besides, the classification predictions (represented as yellow blocks in Fig.~\ref{fig_network}) contain $N$ classification maps, %\yx{ located in separate channels }, 
which model the different semantic parts of objects. %\yx{jointly}.

To achieve the prediction decoupling and collection mentioned in Section~\ref{Prediciton Decoupling}, 
as shown in Fig.~\ref{fig_network}, in parallel with the regression branch, we devise a two-step dynamic point generation module to produce the dynamic boundary points and semantic points at each grid as in Section~\ref{Dynamic_points}. %we devise a single-layer convolution to predict the coarse object boxes and offsets at each grid, producing the dynamic boundary points and semantic points as in Section~\ref{Dynamic_points}. %at each spatial location. 
These two kinds of points are optimized to %\sout{be close to} 
approach the edges or semantic regions of the target object. %\sout{to be detected}.
After having the densely predicted classification and regression maps, we perform prediction collection guided by these dynamic points, where bilinear interpolation is used 
to approximate the collected predictions. 
For the regression maps of multiple scale levels, we use the dynamic boundary points to 
collect the positional offset prediction for each %\yx{box} 
side, with which to produce the final object bounding box. For the classification maps, we incorporate the scores at the dynamic semantic points to jointly identify the object.
The overall dense detection results are the combination of bounding boxes and classification scores produced by dynamic point sets at all grid locations. 
Through prediction decoupling, our proposed network effectively reuses the dense predictions for classification and localization, %obtains localization and classification results at proper positions,
thereby achieving accurate and efficient detection.

\subsection{Training and Inference}

\subsubsection{Training}\label{seq:training}
Our detection network is trained with the following loss:
\begin{equation}
L = L_{cls} + \lambda_1 L_{reg} + \lambda_2 L_{reg_2}      \label{eq_loss}
\end{equation}
where $L_{cls}$ and $L_{reg}$ are the standard classification and regression losses to supervise the final detection results\revised{~\cite{zhang2020bridging,tian2019fcos,lin2017focal}}, 
and $L_{reg_2}$ is an additional regression loss (with the same form of $L_{reg}$) to supervise the learning of coarse object boxes used for dynamic point generation. %predicted by the new branch.
$\lambda_1$ and $\lambda_2$ are hyper-parameters to balance these losses during training. %\sout{The classification loss and regression loss are implemented by focal loss~\cite{lin2017focal} and GIoU loss~\cite{rezatofighi2019generalized}, respectively.}
In our implementation, focal loss~\cite{lin2017focal} is adopted for the classification loss $L_{cls}$, while GIoU loss~\cite{rezatofighi2019generalized} is used for the regression losses $L_{reg}$ and $L_{reg_2}$.

\revised{In the loss $L_{reg_2}$, we use the ground-truth bounding box to supervise the coarse object box estimation. 
Specifically, for each ground-truth bounding box $B^{*}_{i}$ ($1\!\leq\! i\!\leq\!M$) of the current batch, we match it with the coarse object box $B^{'}_{i}$ predicted from the feature map grid closest to the ground-truth bounding box's center. Then, we compute $L_{reg_2}$ by measuring the differences between the ground-truth bounding boxes and their matched coarse boxes with GIoU loss~\cite{rezatofighi2019generalized}:}
\begin{equation}
L_{reg_2} = \frac{1}{M} \sum_{i=1}^{M} \mathrm{GIoU}(B^{*}_{i}, B^{'}_{i}). 
\end{equation}

To compute the classification loss $L_{cls}$ and regression loss $L_{reg}$, we first find the coarse box predictions with IoU larger than 0.6 \wrt the nearest ground-truth bounding box, and assign the dynamic points associated with these coarse boxes as positive samples for different targets, \ie, %\sout{class label and object boundary}
\revised{the object category and the bounding box boundaries}. Then, we take the corresponding ground-truth labels to guide the classification and localization predictions from these dynamic points as well as the position learning for these dynamic points.

\subsubsection{Inference}
During inference, the detection network first densely predicts the classification and regression maps from each level of the feature pyramids. 
Then two sets of dynamic points are generated for each grid location to collect predictions and produce the final classification scores and object bounding boxes. 
Finally, the non-maximum suppression (NMS) with IoU threshold 0.6 is used to determine the final detection results.

\section{Experiments}
The experiments are conducted on the challenging MS COCO 2017 benchmark~\cite{lin2014microsoft}. 
We train the detection model on the \texttt{train2017} split and %\sout{conduct the ablation study on the val2017 set}
evaluate our model on the \texttt{val2017} split. 
We also compare with other methods on the \texttt{test-dev} split, which is the official test set %split
without public ground-truth labels for benchmarking purpose.

\vspace{-0.2cm}

\subsection{Implementation Details}
Following the common experimental conventions~\cite{lin2017focal,tian2019fcos,zhang2020bridging}, we use ResNet-50~\cite{he2016deep} with FPN~\cite{lin2017feature} as the backbone in most of our experiments except when \revised{compared} with other cutting-edge methods. 
The \revised{ResNet-50} has been pre-trained on the ImageNet dataset~\cite{deng2009imagenet}. 
Our detection model is trained with the synchronized stochastic gradient descent (SGD) %\yx{optimizer}
on 4 GPUs with 16 images per minibatch. %\sout{a total of 16 images per minibatch. }\sout{with batch size of 16 images} 
The training procedure lasts for 90k iterations with an initial learning rate of 0.01, which decays by a factor of 10 after 60k iterations and 80k iterations, respectively. The input images are resized to make the shorter edges equal to 800 and the longer sides no larger than 1333. \revised{These hyper-parameters for training follow the previous works~\cite{lin2017focal,tian2019fcos,zhang2020bridging} for a fair comparison.} Besides, only random horizontal image flipping is used in data augmentation. 
Moreover, for Eq.~\eqref{eq_loss}, we set $\lambda_1=2.0$ and $\lambda_2=0.5$. Unless otherwise specified, we adopt $N=9$ in generating the dynamic semantic points. %to model different semantic parts of the object.

\vspace{-0.2cm}

\subsection{Ablation Study}

\subsubsection{Prediction Decoupling}

\begin{table}[!t]
    \renewcommand{\arraystretch}{1.168}
	\caption{The ablation studies of the prediction decoupling. The $D_{loc}$ and $D_{cls}$ refer to applying the prediction decoupling on the localization and classification branches, respectively.
	}
	\begin{center}
	\resizebox{.99988\columnwidth}{!}{
	\begin{tabular}{l|cc|c|cc|ccc}%{l|cc|c|c|c|c|c|c}
		\hline
		Method & $D_{loc}$ & $D_{cls}$ & $\mathrm{AP}$ & $\mathrm{AP}_{50}$ & $\mathrm{AP}_{75}$ & $\mathrm{AP}_{S}$ & $\mathrm{AP}_{M}$ & $\mathrm{AP}_{L}$ \\
		\hline%\hline
		FCOS~\cite{tian2019fcos}    &   &               & 38.6 & 57.4 & 41.4 & 22.3 & 42.5 & 49.8 \\
		ATSS~\cite{zhang2020bridging}    &   &               & 39.3 & 57.5 & 42.8 & 24.3 & 43.3 & 51.3 \\
		\hline%\hline
		\textit{\footnotesize Ours:} & & & & & & & &\\
		PDNet   &           &               & 39.5 & 57.5 & 43.0 & 22.3 & 43.5 & 52.0 \\
		PDNet   &\checkmark &               & 40.8 & 58.4 & 43.6 & 23.5 & 45.0 & 53.5 \\ %with multi-level
		PDNet   &           & \checkmark    & 40.6 & 59.1 & 44.2 & 23.4 & 44.3 & 53.0 \\
		PDNet   &\checkmark & \checkmark    &\textbf{41.8} &\textbf{60.0} &\textbf{45.1} &\textbf{24.7} &\textbf{45.8} &\textbf{55.2} \\ %with multi-level
		\hline
	\end{tabular}
    	}
	\end{center}
	\label{table_pred_decouple}
	\vspace{-0.4cm}
\end{table}

To demonstrate the effectiveness of our prediction decoupling mechanism for accurate detection, we conduct a thorough ablation study. 
\revised{The third row of Table~\ref{table_pred_decouple} shows our baseline without the prediction decoupling mechanism, where the detector follows a prediction manner similar to the previous one-stage methods ATSS~\cite{zhang2020bridging} and FCOS~\cite{tian2019fcos}. This baseline is trained with our positive sample assignment strategy (mentioned in Section~III-E1) and achieves 39.5 AP, which is similar to ATSS's and better than FCOS's (listed in the first two rows of Table~\ref{table_pred_decouple}).}
We first individually add the prediction decoupling to the localization or classification branches and find the performance is improved by 1.3 AP and 1.1 AP respectively, as shown in the 4th and 5th rows of Table~\ref{table_pred_decouple}. 
Furthermore, the last row of Table~\ref{table_pred_decouple} shows the detection performance of our model with the prediction decoupling applied to both the localization and classification, which improves the baseline from 39.5 AP to 41.8 AP (+ 2.3 AP) and achieves the best results among all these ablation variants.

\subsubsection{Points for localization}\label{experiment_loc_pt}

\begin{figure}[!t]
    \setlength{\abovecaptionskip}{1pt}
	\centering
	%\begin{center}
	\includegraphics[width=0.9668\columnwidth, trim=0 0 0 0,clip]{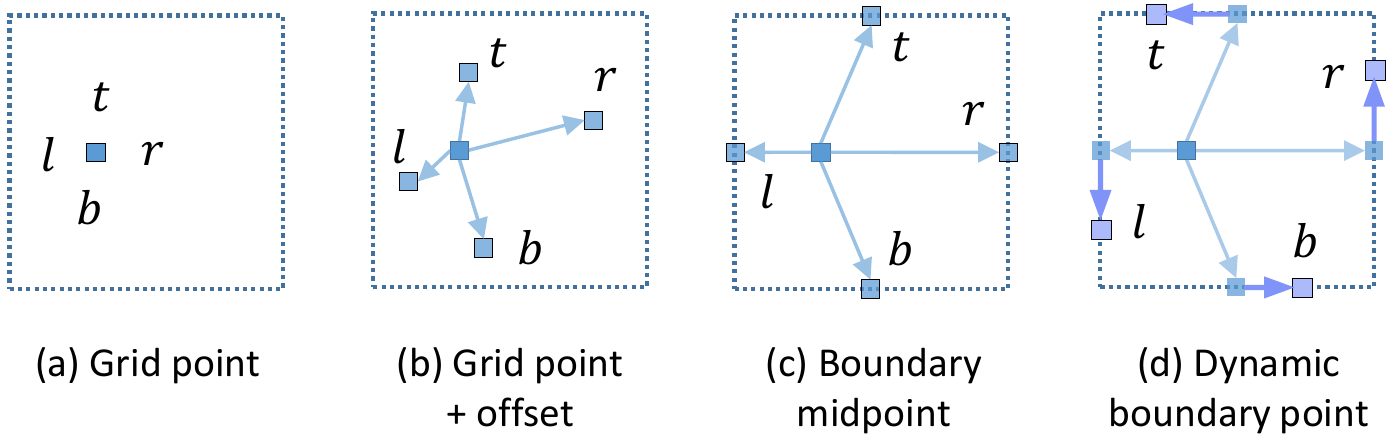} 
	\caption{Different point configurations for localizing each side of the object bounding box. (The dotted rectangle denotes the estimated coarse object box, which provides the prior positions for the dynamic boundary point generation. \revised{The abbreviations $l, t, r, b$ denote which side each point is in charge of.})}
	%\end{center}
	\label{fig_reg_point}
	\vspace{-0.28cm}
\end{figure}

\begin{table}[!t]
\renewcommand{\arraystretch}{1.268}
\caption{Comparison of different point positions used to localize each side of the final object bounding box.}
\begin{center}
\resizebox{.918\columnwidth}{!}{
\begin{tabular}{l|c|ccc}
\hline
Points for localization & $\mathrm{AP}$ & $\mathrm{AP}_{50}$ & $\mathrm{AP}_{75}$ & $\mathrm{AP}_{90}$ \\
\hline
(a)\, Grid point 		& 40.6 & 59.1 & 44.2 & 17.0 \\ 
\hline
(b)\, Grid point $+$ offset 		& 40.8 & 59.4 & 44.4 & 17.8 \\ 
\hline
(c)\, Boundary midpoint  		& 41.0 & 59.7 & 44.6 & 18.0 \\ 
\hline
\textcolor{black}{(d)}\, Dynamic boundary point      & \textbf{41.4} & 59.9  &\textbf{44.8}  & 18.8  \\ 
\textcolor{white}{(}--\textcolor{white}{)}\, 
Dynamic boundary point$\times 2$    & 41.4   & 59.8          & 44.7    & \textbf{19.0}  \\
\textcolor{white}{(}--\textcolor{white}{)}\,
Dynamic boundary point$\times 3$    & 41.3   & \textbf{60.0} & 44.7    & 18.4  \\
\hline
\end{tabular}
}
\end{center}
\label{table_reg_point}
\vspace{-0.4cm}
\end{table}

\revised{
To demonstrate the effectiveness of our dynamic boundary points, in Table~\ref{table_reg_point}, we compare the detection performance when the localization prediction (for each bounding box edge) is collected from different positions. %different configurations of point positions for localization predictions for each side of the object bounding box.
Specifically, based on our PDNet model, we apply different point configurations for prediction decoupling to regress each bounding box edge. 
Fig.~\ref{fig_reg_point} shows the different point configurations: } %for localizing each side of the object bounding box: } 
(a) The original grid point.  % Use the original grid point location.
(b) \revised{An estimated point} from the grid (with the positional offset predicted by the network). %\sout{A point generated}
(c) The midpoint on each boundary of the estimated coarse object box. %(b) Use the boundary center point on each side of the coarse object box estimated.
(d) Our proposed dynamic boundary point. %on each side of the coarse object box.
%(d) Use multiple dynamic boundary points. 
As shown in Table~\ref{table_reg_point}, learning an offset from the original grid improves the AP by 0.2, while the variant with the boundary \revised{midpoint} achieves 0.4 AP higher than that with the grid point, indicating that better localization results can be obtained near the object boundaries.
By introducing the two-step dynamic point generation strategy (\ie, estimating the coarse boundary midpoints first and then shifting them with the predicted offsets), our proposed dynamic boundary points can further improve the performance to 41.4 AP and consistently boost the AP of various IoU metrics, %\yxn{accuracy has its specific meanings in classification} 
which testifies that the two-step generation can better model the object edges. 
The significant improvement of $\mathrm{AP}_{90}$ (+ 1.8 points) shows the great advantage of our method in high-quality localization.  
\revised{We also evaluate different numbers of dynamic boundary points used for localizing each side of the object bounding box. However, as shown in Table~\ref{table_reg_point}, when further increasing the points, the performance varies by only 0.1 AP, showing no significant improvement. 
It indicates that a single dynamic boundary point is sufficient to model each edge of the object bounding box well.}

\begin{figure}[!t]
    \setlength{\abovecaptionskip}{1pt}
	\centering
	%\begin{center}
	\includegraphics[width=0.9986\columnwidth, trim=0 0 0 0,clip]{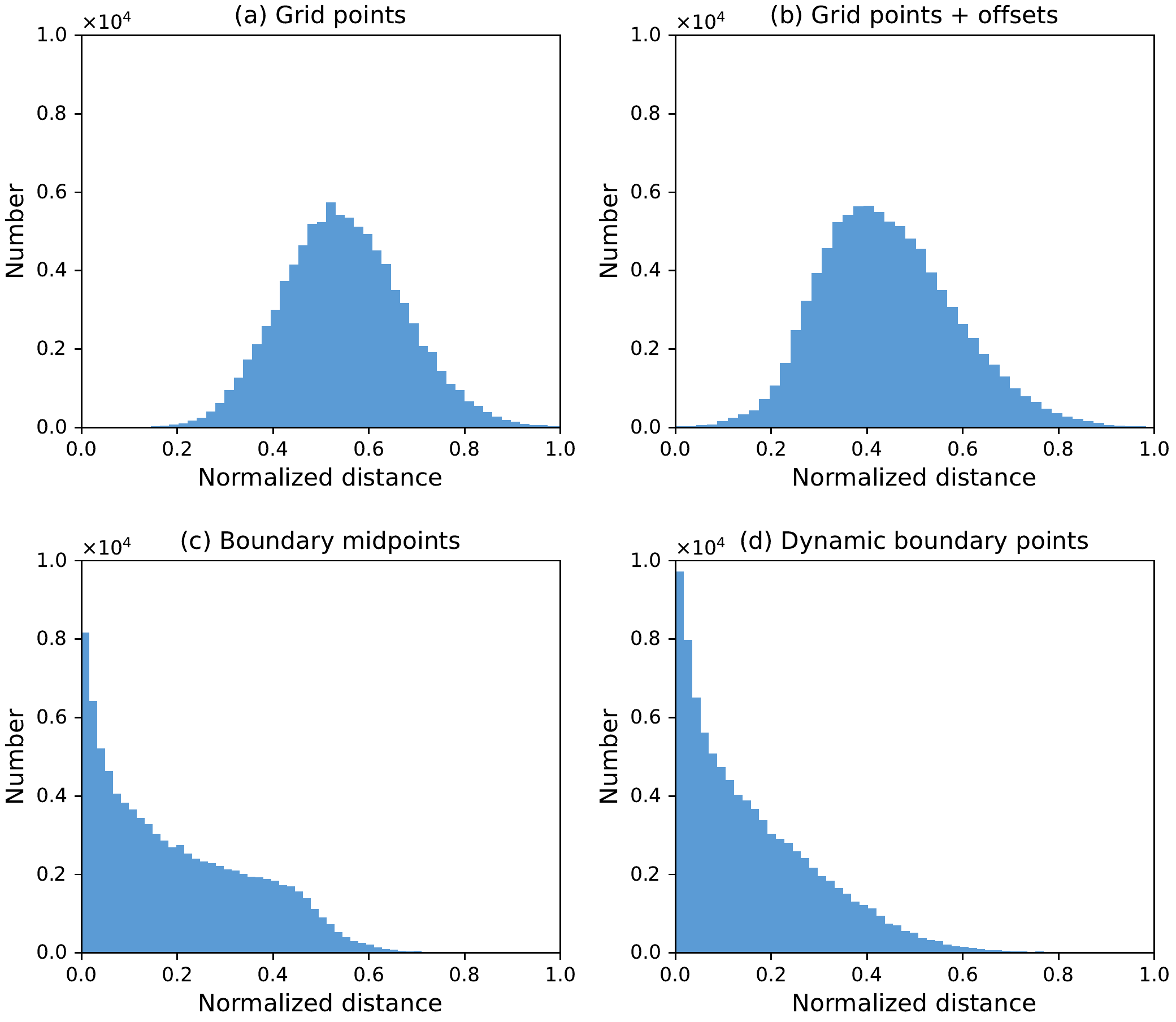} % Reduce the figure size so that it is slightly narrower than the column.
	\caption{The histograms that show the distributions of the normalized distances from different point sets to the object boundaries.} 
	%\end{center}
	\label{fig_reg_point_distribution}
	\vspace{-0.1cm}
\end{figure}

\revised{Moreover, we analyze these different point configurations in Table~\ref{table_reg_point} by measuring their distances to the object boundaries. For this analysis, we first use the ground-truth segmentation masks of objects to obtain their leftmost, topmost, rightmost, and bottommost boundaries. Then, we compute the distances between these outermost object boundaries and the point positions (finally used for object localization) with different configuration variants. Here, we normalize the distances with the ground-truth bounding box sizes. 
Fig.~\ref{fig_reg_point_distribution} plots the distributions of the normalized distances. It can be found that ``(a) grid points'' and ``(b) grid points + offsets'' usually have large distribution densities in the regions relatively distant from the object boundaries. Their accuracies reported in Table~\ref{table_reg_point} are also relatively lower. %As show in Fig.~\ref{fig_reg_point_distribution}, ``(a) grid points'' and ``(b) grid points + offsets'' are located farther from the object boundaries, and their accuracies are also lower. 
The ``(c) boundary midpoints'' are closer to the object boundaries, which improves the performance to 41.0 AP. 
%By contrast, 
With the estimated offsets along the coarse box edges, our proposed ``(d) dynamic boundary points'' further narrow down the distances to the object boundaries and achieve the best performance of 41.4 AP. 
These analyses further demonstrate the necessity of using positions near object boundaries for localization prediction. 
}

\subsubsection{Points for classification}

\begin{figure}[!t]
    \setlength{\abovecaptionskip}{1pt}
	\centering
	\includegraphics[width=0.7468\columnwidth, trim=0 0 0 0,clip]{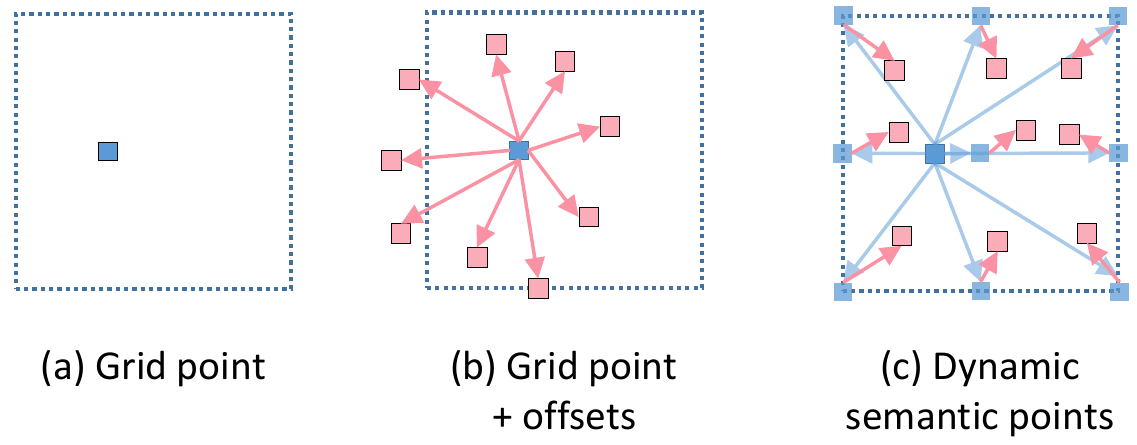} 
	\vspace{-0.1cm}
	\caption{Different point positions for classification prediction collection. (The coarse object box drawn by dotted lines provides the prior positions for dynamic semantic points in the two-step generation.) % \sout{learning}(The dotted rectangle denotes the estimated coarse object box, which provides prior positions for dynamic semantic point learning.)
	}
	\label{fig_cls_point}
	\vspace{-0.26cm}
\end{figure}

\begin{table}[!t]
\renewcommand{\arraystretch}{1.128}
\caption{Comparison of different point positions used for collecting classification predictions.}
\begin{center}
\resizebox{0.918\columnwidth}{!}{
\begin{tabular}{l|c|c|cc|c}
	\hline
Points for classification & $N$ & $\mathrm{AP}$ & $\mathrm{AP}_{50}$ & $\mathrm{AP}_{75}$ & Test\,time\,(ms) \\
\hline %---------------------------------------------------------------
(a)\, Grid point 	        & 1     & 40.4 & 58.5 & 43.5 & 60.3 \\%& 24.4 & 44.3 & 53.0
\hline %---------------------------------------------------------------
\textcolor{white}{(b)}\, Grid point $+$ offsets 	& 2     & 40.8 & 59.1 & 44.1    & 60.6 \\%& 0.2400, 0.4463, 0.5383
\textcolor{white}{(b)}\, Grid point $+$ offsets  & 3     & 41.0 & 59.1 & 44.5       & 60.9\\%, 0.2403, 0.4455, 0.5402
\textcolor{black}{(b)}\, Grid point $+$ offsets  & 5     & 41.1 & 59.5 & 44.4       & 61.2\\%& 24.3 & 44.8 & 0.5428
\textcolor{white}{(b)}\, Grid point $+$ offsets  & 9     & 41.2 & 59.5 & 44.7       & 61.7\\% 0.2392, 0.4476, 0.5425
\textcolor{white}{(b)}\, Grid point $+$ offsets  & 13    & 41.2  & 59.7 & 44.7      & 62.3\\
\textcolor{white}{(b)}\, Grid point $+$ offsets  & 16    & 41.2  & 59.7 & 44.8      & 62.8\\
\hline %---------------------------------------------------------------
\textcolor{white}{(c)}\, Dynamic semantic points & 5     & 41.3  & 59.4  & \textbf{45.1} & 61.2\\
\textcolor{black}{(c)}\, Dynamic semantic points & 9     &\textbf{41.4}  & \textbf{59.9} & 44.8 & 61.7 \\ %& 24.8 & 45.3
\textcolor{white}{(c)}\, Dynamic semantic points & 13     & 41.4 & 59.7  & 44.8 & 62.3 \\ 
\textcolor{white}{(c)}\, Dynamic semantic points & 16     & 41.4 & 59.8  & 44.9 & 62.8 \\ 
\hline
\end{tabular}
}
\end{center}
\label{table_cls_point}
\vspace{-0.36cm}
\end{table}

\begin{table}[!t]
\renewcommand{\arraystretch}{1.148}
\caption{Comparison of different levels for prediction collection.}
\begin{center}
\resizebox{.99988\columnwidth}{!}{
\begin{tabular}{l|l|c|ccc|ccc}
    \hline
            %\multicolumn{2}{c|}{Levels $\mathbb{N}(s_0)$}
            & Levels $\mathbb{N}(s_0)$
                            &$\mathrm{AP}$ & $\mathrm{AP}_{50}$ & $\mathrm{AP}_{70}$ & $\mathrm{AP}_{90}$ & $\mathrm{AP}_{S}$ & $\mathrm{AP}_M$ & $\mathrm{AP}_L$ \\
	\hline
	\multirow{5}{*}{Loc.}
	        &$[s_0]$ 		            & 41.4  & 59.9  & 49.9  & 18.8  & 24.8  & 45.3  & 54.7 \\
	        &$[s_0\!-\!1]$ 		        & 41.6  & 59.7  & 49.5  & 19.8  &\textbf{25.3}  & 45.5  & 54.7\\
	        &$[s_0\!-\!1, s_0]$         &\textbf{41.8} &\textbf{60.0} & 50.1 &\textbf{20.2} & 24.7  &\textbf{45.8}  &\textbf{55.2}\\
	        &$[s_0, s_0\!+\!1]$         & 41.3  & 59.9 & 49.9   & 18.2 & 24.4 & 45.1  & 54.6\\ 
	        &$[s_0\!-\!1, s_0, s_0\!+\!1]$   & 41.8     & 59.8 & \textbf{50.2}   & 19.4  & 24.9  & 45.8  & 54.8\\
	        &$[s_0\!-\!2, s_0\!-\!1,s_0]$   & 31.5  & 43.3 & 37.4   & 17.2  & 4.6  & 45.8  & 55.1\\
	\hline
    \multirow{4}{*}{Cls.}
		    &$[s_0]$                        & 41.4 & 59.9 & 49.9 & 18.8 & 24.8  & 45.3  & 54.7  \\
		    &$[s_0\!-\!1, s_0]$             & 41.2 & 59.6 & 49.7 & 18.8 & 24.2  & 44.8  & 54.0  \\ 
		    &$[s_0, s_0\!+\!1]$             & 41.2 & 59.5 & 49.8 & 18.5 & 25.2  & 44.9  & 54.2  \\
		    &$[s_0\!-\!1, s_0, s_0\!+\!1]$  & 41.3 & 60.0 & 49.9 & 18.5 & 25.0  & 45.0  & 54.0  \\
	\hline
\end{tabular}
}
\end{center}
\label{table_pred_level}
\vspace{-0.36cm}
\end{table}

\begin{table}[!t]
    \renewcommand{\arraystretch}{1.12}
    \caption{Comparison of the aggregate functions.}
	\begin{center}
	\resizebox{.988\columnwidth}{!}{
		\begin{tabular}{l|c|c|cc|ccc}
	    \hline
        %\multicolumn{2}{c|}{Aggregate function}  
        & Aggregate function & $\mathrm{AP}$ & $\mathrm{AP}_{50}$ & $\mathrm{AP}_{75}$ & $\mathrm{AP}_{S}$ & $\mathrm{AP}_M$ & $\mathrm{AP}_L$ \\
		\hline
		\multirow{2}{*}{Loc.}
		                            & Avg. w/o weights   & 41.6  & 59.9  & 44.9  & 25.0  & 45.4  & 54.6  \\
		                            & Avg. w/ weights     & 41.8  & 60.0  & 45.1  & 24.7  & 45.8  & 55.2  \\
		\hline
		\multirow{2}{*}{Cls.} 
		                            & Sum w/o weights   & 41.4  & 59.9  & 44.8  & 24.8  & 45.3  & 54.7 \\ %
                                    & Sum w/ weights     & 41.3  & 59.7  & 44.9  & 24.7  & 44.5  & 54.4 \\ %
		\hline
	    \end{tabular}
	}
	\end{center}
	\label{table_pred_aggre}
	\vspace{-0.28cm}
\end{table}

\begin{table*}[!t]
\renewcommand{\arraystretch}{1.1}
\caption{Comparison of our method with other state-of-the-art detectors on \revised{the} MS COCO \texttt{test-dev} split. ``$\dag$'' indicates using a wider scale range [480:960] for multi-scale training. }%multi-scale training with a wider scale range [480, 960].}
%\centering
\begin{center}
\resizebox{0.68\textwidth}{!}{
\begin{tabular}{l|c|c|cc|ccc}%{l|c|l|l|l|l|l|l}
\hline
%Method & Backbone & $AP$ & $AP_{50}$ & $AP_{75}$ & $AP_S$ & $AP_M$ & $AP_L$ \\
Method & Backbone & $\mathrm{AP}$ & $\mathrm{AP}_{50}$ & $\mathrm{AP}_{75}$ & $\mathrm{AP}_S$ & $\mathrm{AP}_M$ & $\mathrm{AP}_L$ \\
\hline
FPN~\cite{lin2017feature}           & ResNet-101  & 36.2 & 59.1 & 39.0 & 18.2 & 39.0 & 48.2 \\
Mask R-CNN~\cite{he2017mask}        & ResNet-101  & 38.2 & 60.3 & 41.7 & 20.1 & 41.1 & 50.2 \\
Cascade R-CNN~\cite{cai2018cascade} & ResNet-101  & 42.8 & 62.1 & 46.3 & 23.7 & 45.5 & 55.2 \\
\hline
RetinaNet~\cite{lin2017focal}           & ResNet-101    & 39.1 & 59.1 & 42.3 & 21.8 & 42.7 & 50.2 \\
%FoveaBox~\cite{kong2019foveabox}        & ResNet-101    & 40.6 & 60.1 & 43.5 & 23.3 & 45.2 & 54.5 \\
FoveaBox~\cite{kong2020foveabox}        & ResNet-101    & 40.8 & 61.4 & 44.0 & 24.1 & 45.3 & 53.2 \\
FSAF~\cite{zhu2019feature}              & ResNet-101    & 40.9 & 61.5 & 44.0 & 24.0 & 44.2 & 51.3 \\
FCOS~\cite{tian2019fcos}                & ResNet-101    & 41.5 & 60.7 & 46.3 & 23.7 & 45.5 & 55.2 \\
FreeAnchor~\cite{zhang2019freeanchor}   & ResNet-101    & 43.1 & 62.2 & 46.4 & 24.5 & 46.1 & 54.8 \\
%FreeAnchor~\cite{zhang2019freeanchor}   & ResNeXt-64x4d-101     & 44.9 & 64.3 & 48.5 & 26.8 & 48.3 & 55.9 \\
FreeAnchor$\dag$~\cite{zhang2019freeanchor}   & ResNeXt-64x4d-101     & 46.0 & 65.6 & 49.8 & 27.8 & 49.5 & 57.7 \\
ATSS~\cite{zhang2020bridging}           & ResNet-101    & 43.6 & 62.1 & 47.4 & 26.1 & 47.0 & 53.6 \\
ATSS~\cite{zhang2020bridging}           & ResNeXt-64x4d-101    & 45.6 & 64.6 & 49.7 & 28.5 & 47.0 & 53.6 \\
GFL~\cite{li2020generalized}           & ResNet-101    & 45.0 & 63.7 & 48.9 & 27.2 & 48.8 & 54.5 \\
\hline
% AlignDet~\cite{chen2019revisiting}      & ResNet-101        & 42.0 & 62.4 & 46.5 & 24.6 & 44.8 & 53.3 \\
RepPoints~\cite{yang2019reppoints}      & ResNet-101-DCN    & 45.0 & 66.1 & 49.0 & 26.6 & 48.6 & 57.5 \\
BorderDet~\cite{qiu2020borderdet}       & ResNet-101        & 45.4 & 64.1 & 48.8 & 26.7 & 48.3 & 56.5 \\
BorderDet~\cite{qiu2020borderdet}       & ResNeXt-64x4d-101 & 46.5 & 65.7 & 50.5 & 29.1 & 49.4 & 57.5 \\
RepPoints v2$\dag$~\cite{chen2020reppoints}      & ResNet-101    & 46.0 & 65.3 & 49.5 & 27.4 & 48.9 & 57.3 \\
RepPoints v2$\dag$~\cite{chen2020reppoints}      & ResNeXt-64x4d-101    & 47.8 & 67.3 & 51.7 & 29.3 & 50.7 & 59.5 \\
RepPoints v2$\dag$~\cite{chen2020reppoints}      & ResNeXt-64x4d-101-DCN    & 49.4 & 68.9 & 53.4 & 30.3 & 52.1 & 62.3 \\
\hline
{\scriptsize\textit{Ours:}} & & & & & & &\\
% PDNet                                & ResNet-101 & 45.5 & 64.5 & 49.5 & 27.3 & 48.8 & 56.4 \\
% PDNet$\dag$                                & ResNet-101 & 46.2 & 65.1 & 50.5 & 27.8 & 49.7 & 57.5 \\
% PDNet                                & ResNeXt-64x4d-101 & 47.3 & 66.5 & 51.5 & 29.4 & 50.4 & 58.4 \\
% PDNet*                                & ResNeXt-64x4d-101 & 48.4 & 67.6 & 52.8 & 30.5 & 51.8 & 59.6 \\
PDNet                               & ResNet-50 & 44.3 & 62.9 & 48.0 & 26.5 & 47.6 & 54.9 \\
PDNet$\dag$                         & ResNet-50 & 45.0 & 63.5 & 48.6 & 26.9 & 48.4 & 55.9 \\
PDNet                                & ResNet-101 & 45.7 & 64.5 & 49.7 & 27.6 & 49.2 & 56.7 \\
PDNet$\dag$                                & ResNet-101 & 46.6 & 65.3 & 50.6 & 28.0 & 50.2 & 58.0 \\
PDNet                                & ResNeXt-64x4d-101 & 47.4 & 66.6 & 51.5 & 29.6 & 50.6 & 58.5 \\
% PDNet$\dag$       & ResNeXt-64x4d-101 & \textbf{48.7} & \textbf{67.6} & \textbf{52.9} & \textbf{30.5} & \textbf{52.2} & \textbf{60.2} \\
PDNet$\dag$         & ResNeXt-64x4d-101 & 48.7 & 67.6 & 52.9 & 30.5 & 52.2 & 60.2 \\
PDNet$\dag$         & ResNeXt-64x4d-101-DCN & \textbf{50.1} & \textbf{68.9} & \textbf{54.5} & \textbf{31.4} & \textbf{53.2} & \textbf{62.6} \\
\hline
\end{tabular}
}
\end{center}
\label{table_stoa}
\vspace{-0.2cm}
\end{table*}

\revised{In Table~\ref{table_cls_point}, based on our PDNet model, we compare the detection performance when applying different configurations of point positions to collect the classification predictions.} %\sout{  different point configurations for collecting and aggregating classification predictions}. 
Fig.~\ref{fig_cls_point} presents these different configurations, including the following: (a) The original grid location. (b) A set of divergent points generated from the grid (with $N$ offsets predicted by the network). (c) Our proposed dynamic semantic points.
\revised{We first evaluate the performance when increasing the  point number $N$ for classification prediction collection.} 
As shown in Table~\ref{table_cls_point}, the AP value increases (from 40.4) as more points are employed until it reaches the saturation point of 41.2 AP near $N = 9$. 
This shows that the object category can be better recognized with multiple classification predictions from different semantic parts of the object. 
%\yx{By introducing the proposed two-step dynamic point learning process 
With the dynamic semantic points established by the two-step generation process
%By establishing the dynamic semantic points with the two-step learning process
(\ie, estimating the coarse box first and then the offsets from the prior positions), %\sout{predicting}mentioned in Sec.xxx )
the performance can be further improved to 41.4 AP.
These improvements testify that the two-step dynamic semantic point generation can more easily find various semantic regions of the object and thereby obtain better classification results.
\revised{Moreover, as shown in Table~\ref{table_cls_point}, the model's inference time does not increase much when gathering more classification predictions (\eg, 1.4\,ms\,$\uparrow$ when $N=9$). The efficient prediction collection process allows us to improve accuracy while keeping the efficiency. 
}

\subsubsection{Multi-level prediction collection}

As mentioned in Section~\ref{sec:localization}, we can collect the regression predictions from multiple scale levels $\mathbb{N}(s_0)$ to improve the object localization in level $s_0$. 
Here, we further investigate the performance variation when collecting the predictions from different sets of scale levels $\mathbb{N}(s_0)$ for object localization. 
As shown in the third row of Table~\ref{table_pred_level}, utilizing the predictions from the $[s_0\!-\!1, s_0]$ levels achieves the best performance of 41.8 AP, and the localization accuracy is notably improved with a gain of 1.4 $\mathrm{AP}_{90}$ compared with that only taking the predictions from $[s_0]$. %brings a gain of 1.4 in $\mathrm{AP}_{90}$ for high-quality localization. 
\revised{Introducing the predictions from the $s_0\!+\!1$ level brings no further improvement, as the regression map of the $s_0\!+\!1$ level has a lower resolution, which is not very suitable for object localization of the current scale level $s_0$.} 
When further using the predictions from the $s_0\!-\!2$ level, the performance of small object detection deteriorates (to 4.6 $\mathrm{AP}_{S}$) and therefore results in a much lower overall AP score. 
In Table~\ref{table_pred_level}, we also evaluate the effect of collecting classification predictions from adjacent levels, but no performance improvement is observed, %from the experimental results. 
which may be attributed to that the adjacent feature pyramid levels 
encode similar semantic information and cannot benefit each other anymore.

\subsubsection{Prediction aggregate function}

In our implementation, we use the predicted weights to aggregate the localization predictions, while the classification predictions are aggregated by summation without normalized weights. 
In Table~\ref{table_pred_aggre}, we further compare 
the aggregation function variants that w/ or w/o aggregation weights. 
For the object localization, weighting %calculating the weighted average of 
the predictions from different levels leads to better accuracy, since it allows us to select the suitable levels %\sout{to obtain} 
for more precise localization results. 
However, for the classification predictions, the weighted aggregation does not make much difference in performance, indicating that the object category may need to be %jointly 
determined by the classification predictions from different semantic regions together. 
% without selection. 

\begin{figure*}[!t]
	\centering
	\includegraphics[width=0.998\textwidth]{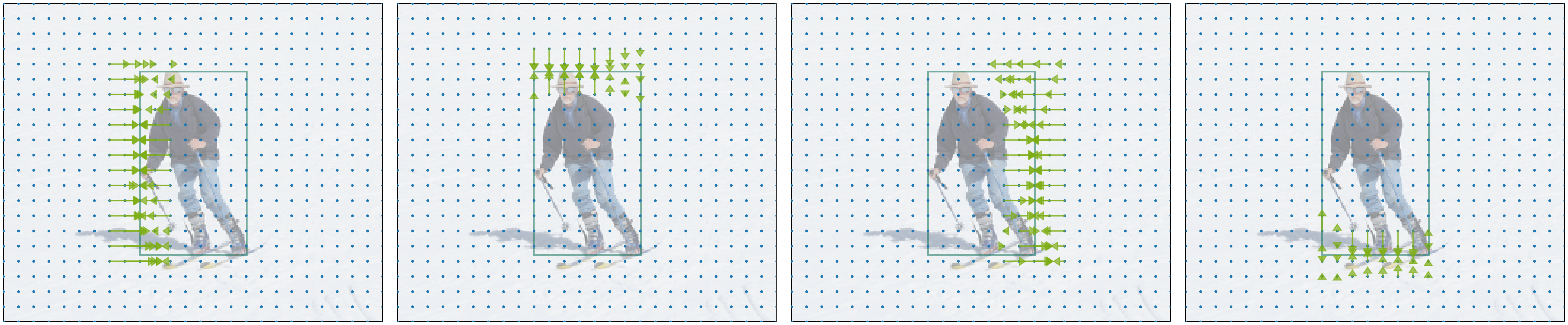}
	\caption{Visualization of the regression maps used to locate the left, top, right, and bottom sides of the object bounding box. For clear demonstration, we only show the predicted offsets in the boundary areas. We can see that the positional offsets from the grids near the object edges accurately match the residual distances to the corresponding bounding box edges. %\sout{point to the} \sout{boundaries of the object bounding box} 
	}
	\label{fig_reg_map}
\end{figure*}

\begin{figure*}[!t]
	\centering
	\includegraphics[width=0.98\textwidth]{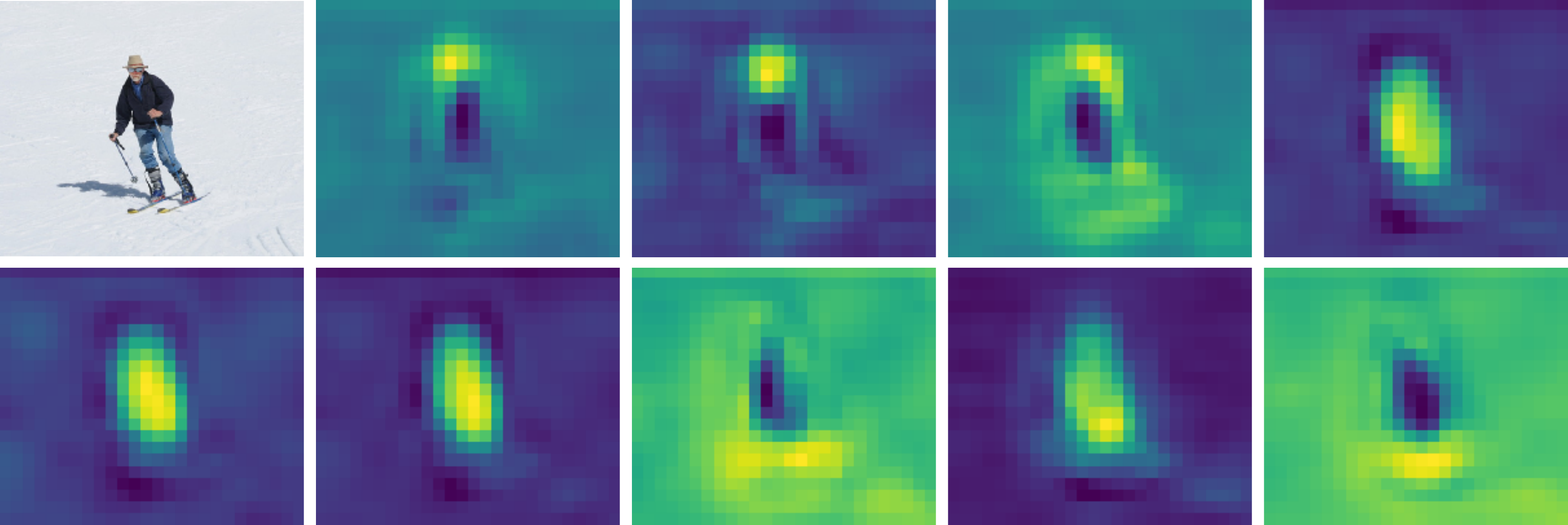}
	\caption{Visualization of the classification maps for the person. %\textit{w.r.t.} the category ``person''. 
	These classification maps produce strong activations in different areas of this person, showing that they model the semantic information of various object regions respectively. }
	%\sout{allowing us to gather predictions from them to jointly identify the category}
	%The classification maps are sensitive to different areas of the person, showing that they model the semantic information of various object regions.}
	\label{fig_cls_map}
	\vspace{-0.2cm}
\end{figure*}

\subsection{Comparisons with State-of-the-art Methods}

We compare our detector PDNet on the \texttt{test-dev} split of \revised{the} MS COCO benchmark with other state-of-the-art methods in Table~\ref{table_stoa}. As in previous works~\cite{lin2017focal,tian2019fcos}, we adopt the multi-scale training strategy by randomly scaling the shorter side of the image to the range from 640 to 800. The training iterations are also doubled to 180k, with the learning rate reduced by 10 times at 120k and 160k iterations, respectively. To compare with the methods that adopt a wider scale range [480:960] for multi-scale training, we also apply this strategy in training our detection model for \revised{a} fair comparison. %accordingly under this same strategy. 
The other settings are kept the same as in the previous experiments.

As shown in Table~\ref{table_stoa}, our detector with the ResNet-101 achieves 45.7 AP without bells and whistles, and outperforms other methods using the same backbone, including FCOS~\cite{tian2019fcos} (41.5 AP), FreeAnchor~\cite{zhang2019freeanchor} (43.1 AP), ATSS~\cite{zhang2020bridging} (43.6 AP), and GFL~\cite{li2020generalized} (45.0 AP, with a wider training scale range [480:800]). Compared with the recently proposed BorderDet~\cite{qiu2020borderdet} and RepPoints~v2~\cite{chen2020reppoints}, our method also performs better with a much simpler network architecture. %\sout{while being} much simpler
With a larger backbone ResNeXt-64x4d-101, we can further improve the AP from 45.7 to 47.4, which is significantly higher than BorderDet~\cite{qiu2020borderdet} (46.5 AP) under the same setting. %This result surpasses all the methods including BorderDet, which shows the great advantage of our method. 
By utilizing a wider training scale range [480:960], our method with a single ResNeXt-64x4d-101 reaches 48.7 AP\revised{, and our model using ResNeXt-64x4d-101-DCN achieves 50.1 AP with single-scale testing, outperforming other methods by an appreciable margin.} %large margin.

\revised{In Table~\ref{table_multi_scale_test}, we further report our multi-scale testing results in comparison with ATSS~\cite{zhang2020bridging} and RepPoints~v2~\cite{chen2020reppoints}. 
RepPoints~v2 follows the multi-scale testing strategy of ATSS~\cite{zhang2020bridging}, where each image is resized to 13 different scales and flipped horizontally for testing, which incurs a high processing cost. %which is very time-consuming. 
For a fair comparison, we follow this testing strategy and achieve 52.3 in AP, higher than both ATSS (50.1) and RepPoints v2 (52.1). While this multi-scale testing brings further improvement, it is particularly time-consuming and may not be suitable for practical usage. }

\subsection{Visualization of the Regression and Classification Maps}
As shown in Fig.~\ref{fig_reg_map}, we visualize the dense offset regression maps used to locate the left, top, right, and bottom edges of the object bounding box. 
For clear illustration, we only show the offset predictions in the areas around the object boundary. It can be observed that the positional offsets estimated at the grids near the object edges %\sout{accurately point to}
accurately match the residual distances to the corresponding boundaries of the object bounding box. 
This provides the foundation for accurate object localization.

In Fig.~\ref{fig_cls_map}, for the same input image, we present the 
predicted classification maps for the person. 
As shown in Fig.~\ref{fig_cls_map}, these classification maps produce strong activations on different parts of this person, \eg, the head, feet, body, \etc. 
This implies that the different classification maps can model the semantic information of different parts of the object, %\sout{and complementary regions}
which allows us to gather predictions from them to jointly identify the object category.

\begin{table}[t]
\renewcommand{\arraystretch}{1.186}
\caption{Comparison with the state-of-the-art methods with single-scale and multi-scale testing.}
\begin{center}
\resizebox{.99988\columnwidth}{!}{
\begin{tabular}{l|c|c|cc|ccc}
\hline
\multicolumn{2}{c|}{Single-scale testing}  & $\mathrm{AP}$ & $\mathrm{AP}_{50}$ & $\mathrm{AP}_{75}$ & $\mathrm{AP}_{S}$ & $\mathrm{AP}_M$ & $\mathrm{AP}_L$ \\ 
\hline
ATSS~\cite{zhang2020bridging}               & ResNeXt-101-DCN   & 47.7  & 66.5  & 51.9  & 29.7  & 50.8  & 59.4  \\
RepPoints\,v2\,\cite{chen2020reppoints}       & ResNeXt-101-DCN     & 49.4  & 68.9  & 53.4  & 30.3  & 52.1  & 62.3  \\
PDNet (ours)    &  ResNeXt-101-DCN  & \textbf{50.1}  & 68.9  & 54.5  & 31.4 & 53.2 & 62.6  \\
\hline
\multicolumn{2}{c|}{Multi-scale testing}  & $\mathrm{AP}$ & $\mathrm{AP}_{50}$ & $\mathrm{AP}_{75}$ & $\mathrm{AP}_{S}$ & $\mathrm{AP}_M$ & $\mathrm{AP}_L$ \\ \hline
ATSS~\cite{zhang2020bridging}               & ResNeXt-101-DCN   & 50.7 & 68.9 & 56.3 & 33.2 & 52.9 & 62.4 \\
RepPoints\,v2\,\cite{chen2020reppoints}       & ResNeXt-101-DCN   & 52.1 & 70.1 & 57.5 & 34.5 & 54.6 & 63.6 \\
PDNet (ours)    &  ResNeXt-101-DCN  & \textbf{52.3}  & 70.1  & 58.0  & 35.3 & 54.4 & 63.3  \\
\hline
\end{tabular}
}
\end{center}
\label{table_multi_scale_test}
\vspace{-0.4cm}
\end{table}

\subsection{Visualization of Detection Results}

\begin{figure*}[t] % trim=l, b, r, t
	\centering
    \begin{minipage}[t]{0.9998\textwidth}
		\centering
		\includegraphics[height=0.24\textwidth, trim=0 0 98 0,clip]{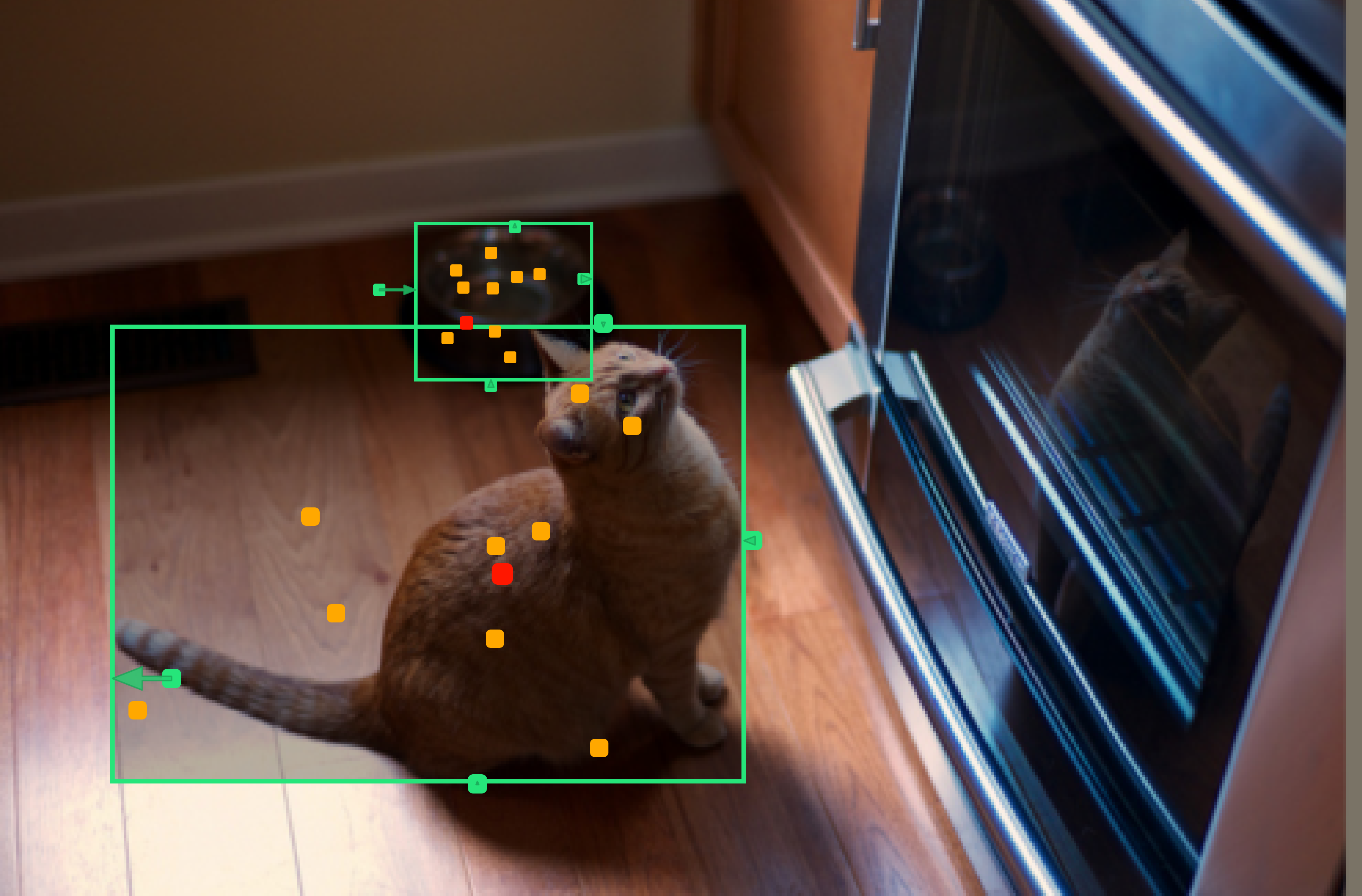}
		\includegraphics[height=0.24\textwidth, trim=0 0 20 0,clip]{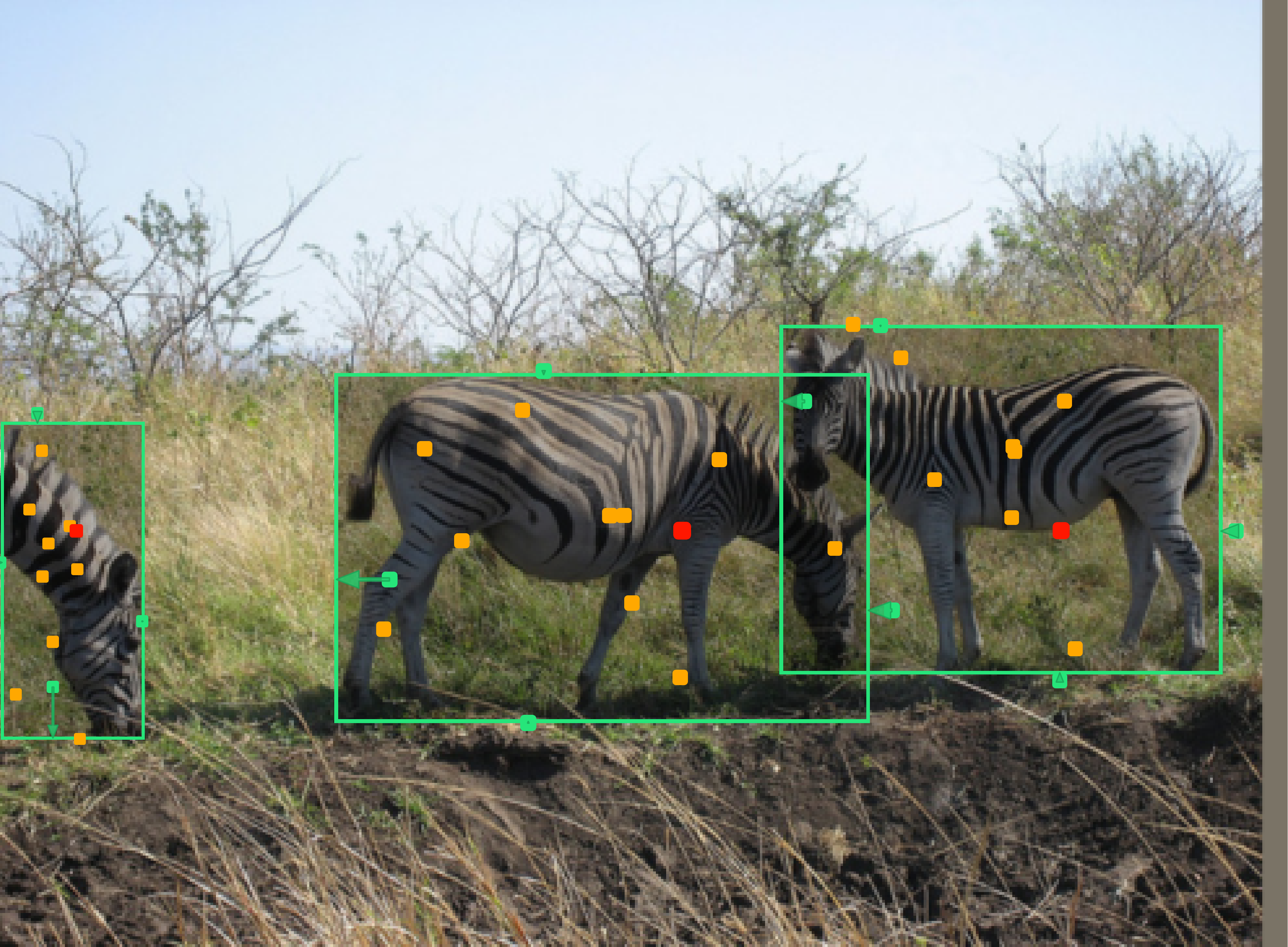}
		\includegraphics[height=0.24\textwidth, trim=28 42 0 0,clip]{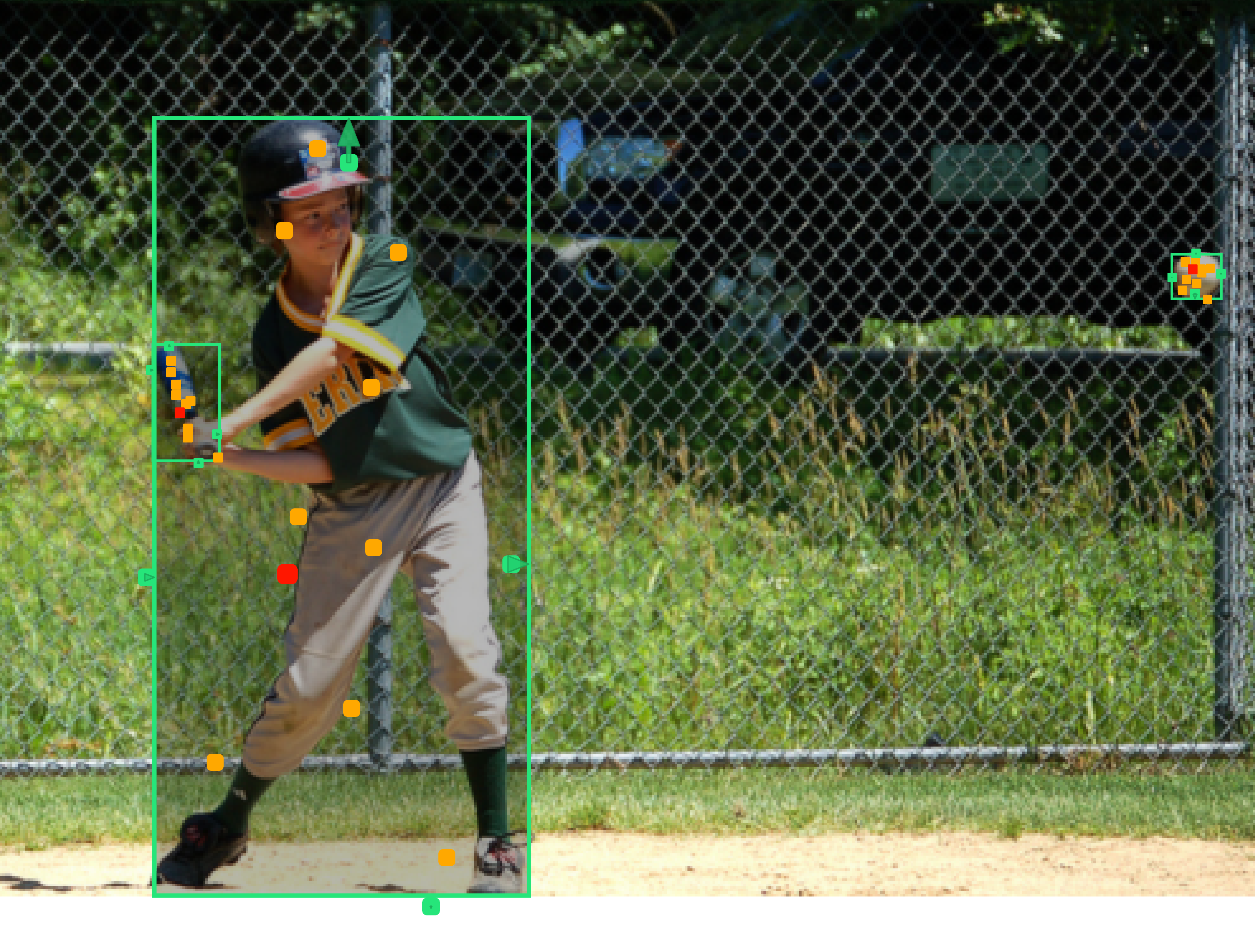}
	\end{minipage}%
	\vspace{0.05cm}
	\begin{minipage}[t]{0.998\textwidth}
		\centering
		\includegraphics[height=0.268\textwidth, trim=20 20 0 0,clip]{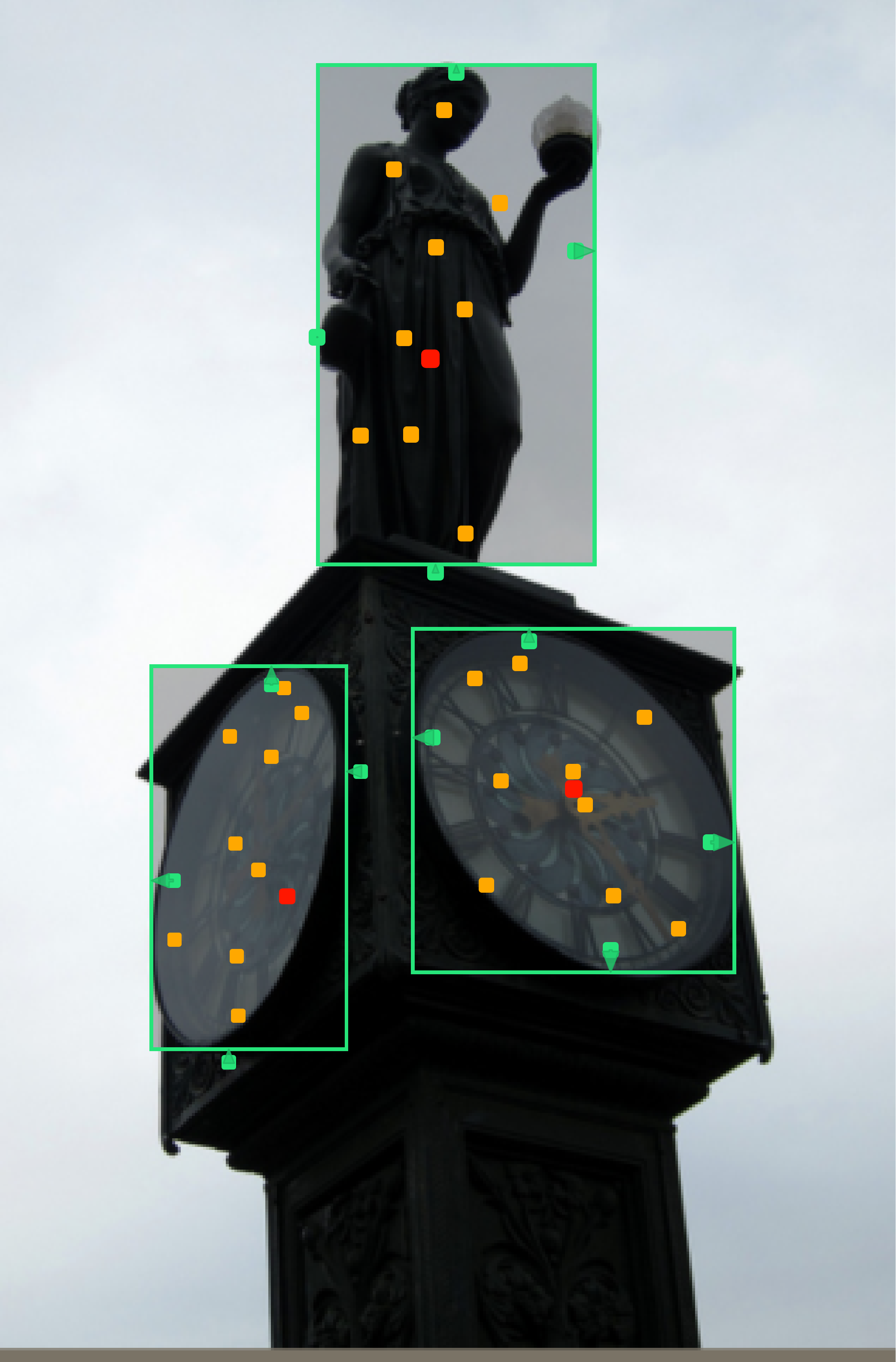}
		\includegraphics[height=0.268\textwidth, trim=0 0 20 0,clip]{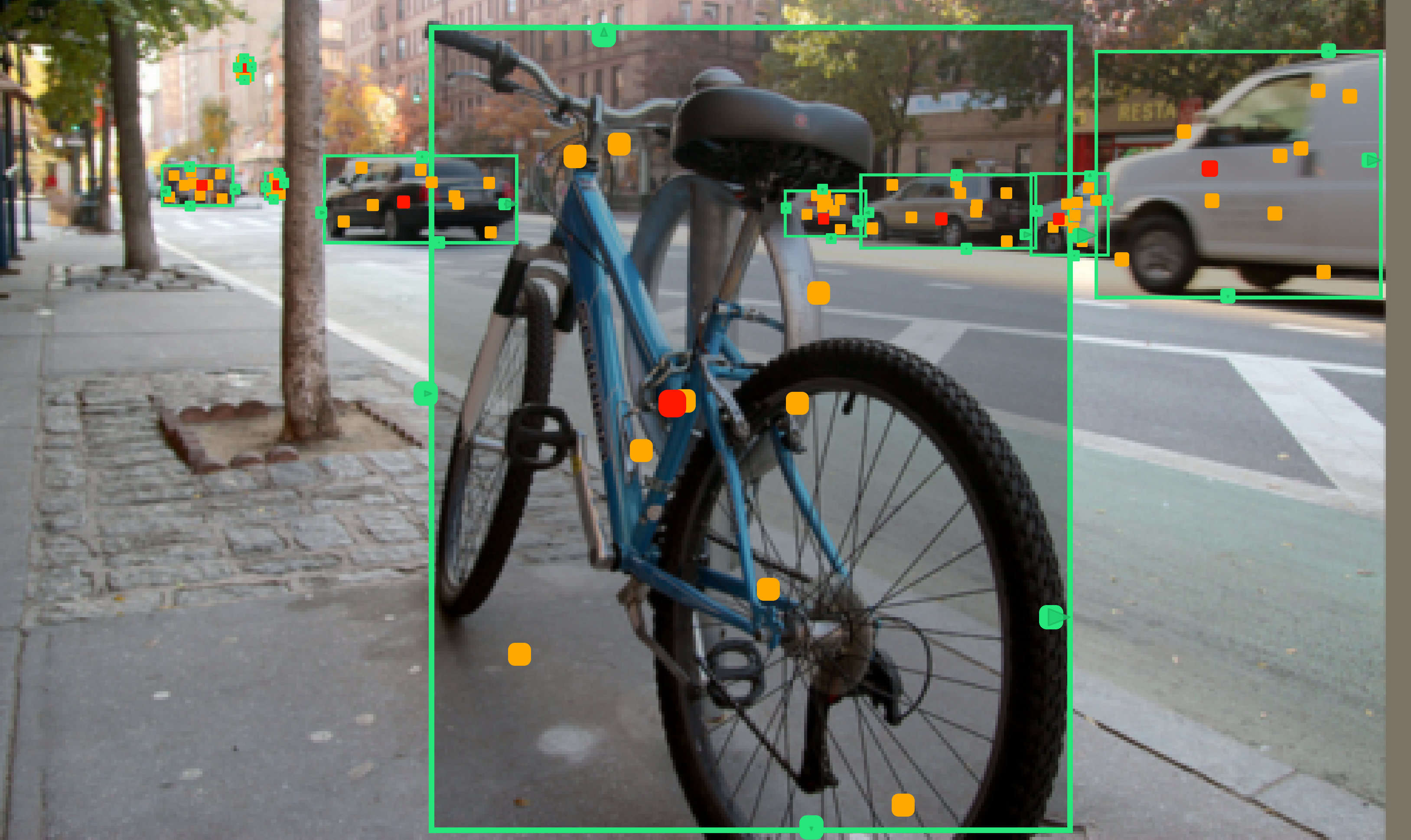}
		\includegraphics[height=0.268\textwidth, trim=0 0 20 0,clip]{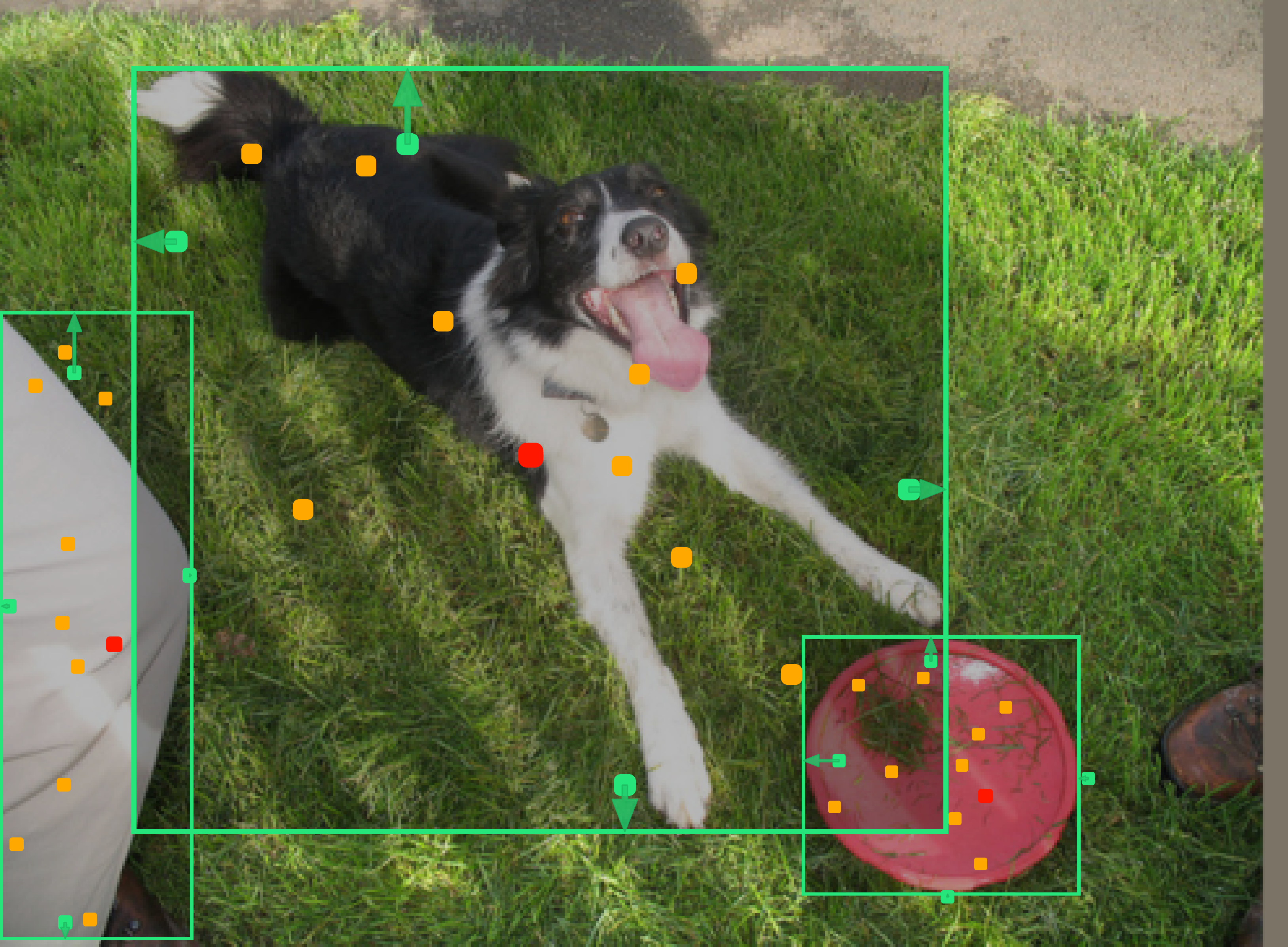}
	\end{minipage}%
	\vspace{0.05cm}
	\\
	\begin{minipage}[t]{0.998\textwidth}
		\centering
		\includegraphics[height=0.228\textwidth, trim=0 0 20 6,clip]{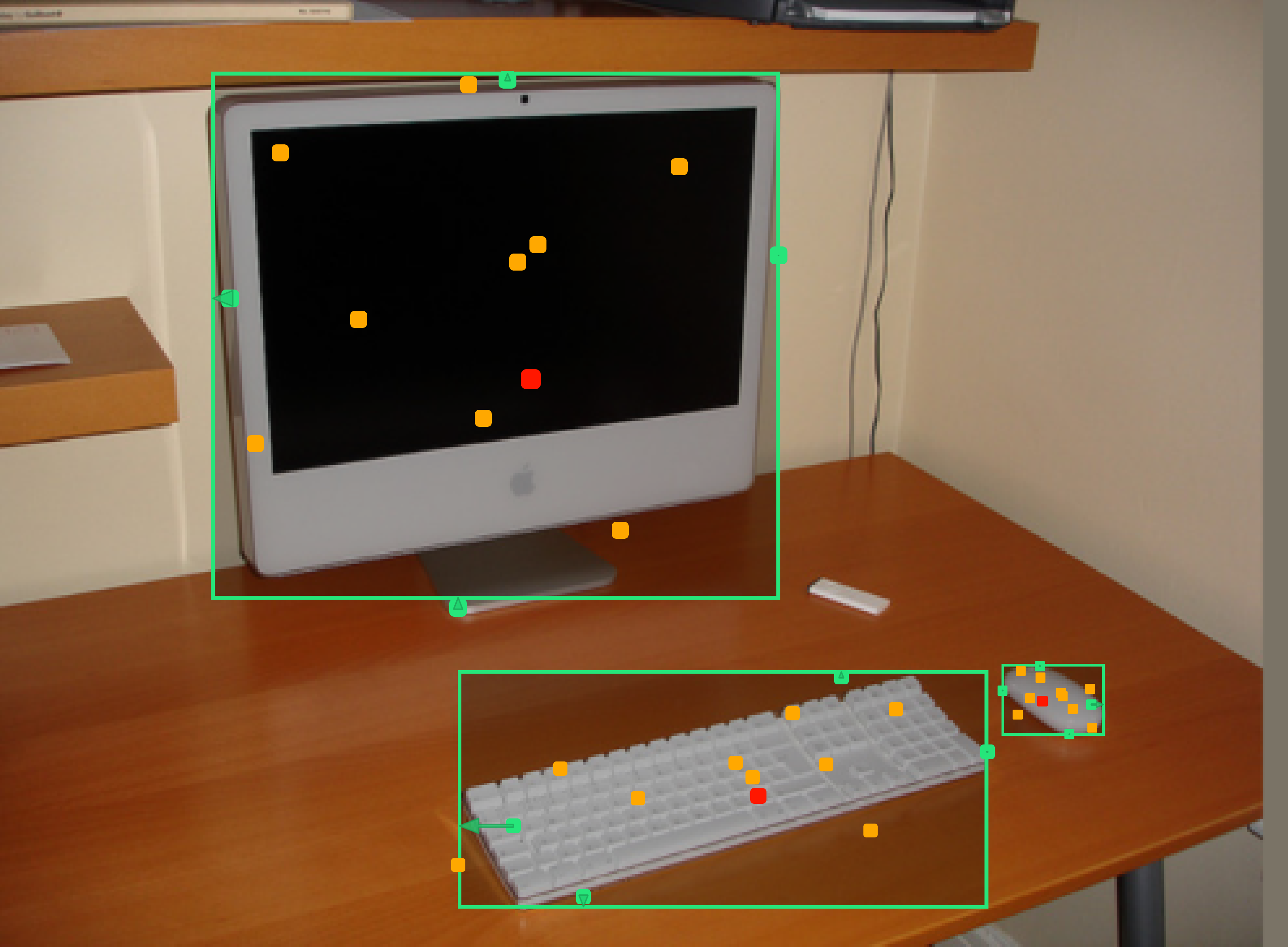}
		\includegraphics[height=0.228\textwidth, trim=0 0 20 18,clip]{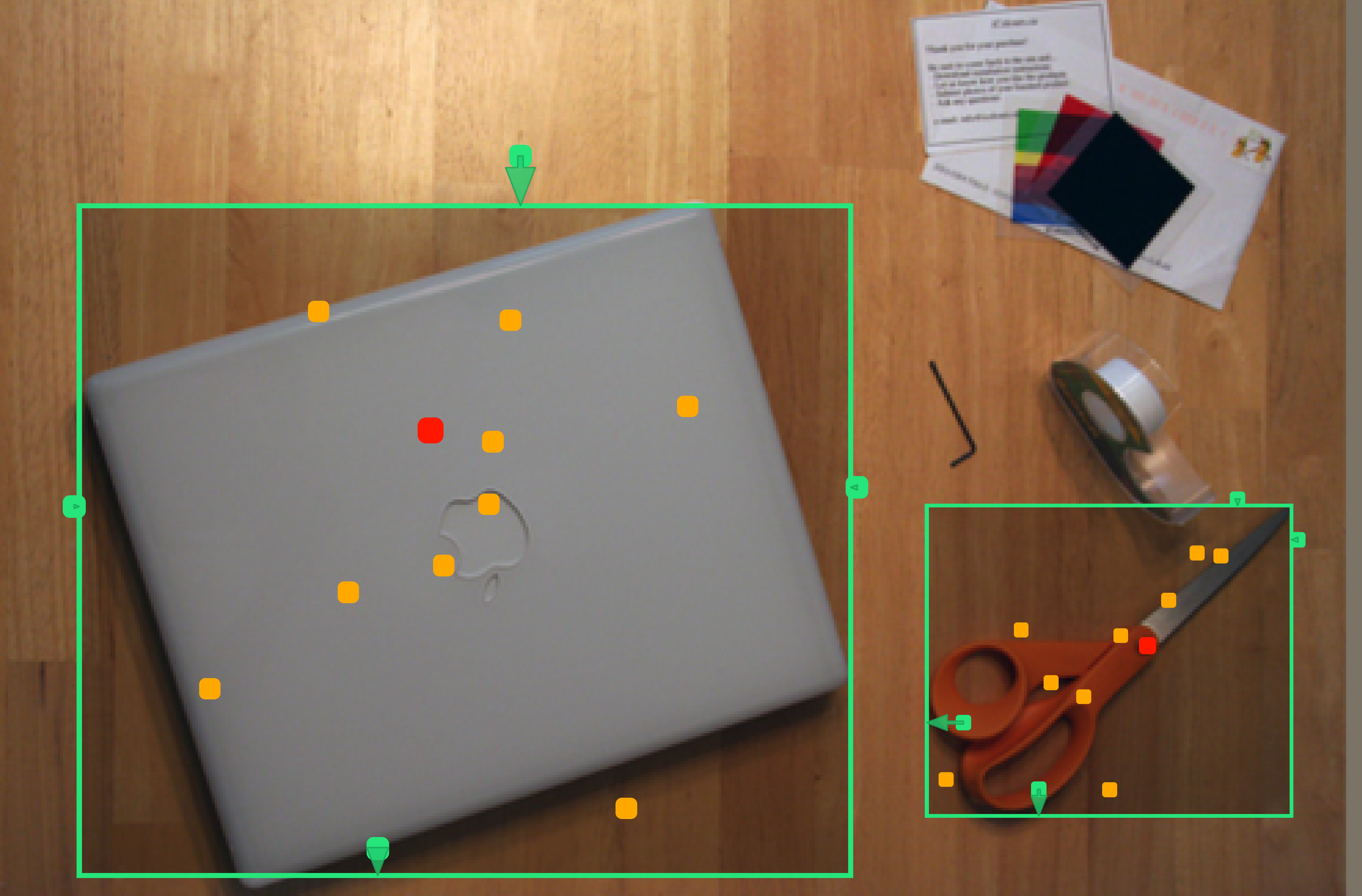}
	\includegraphics[height=0.228\textwidth, trim=0 18 20 0,clip]{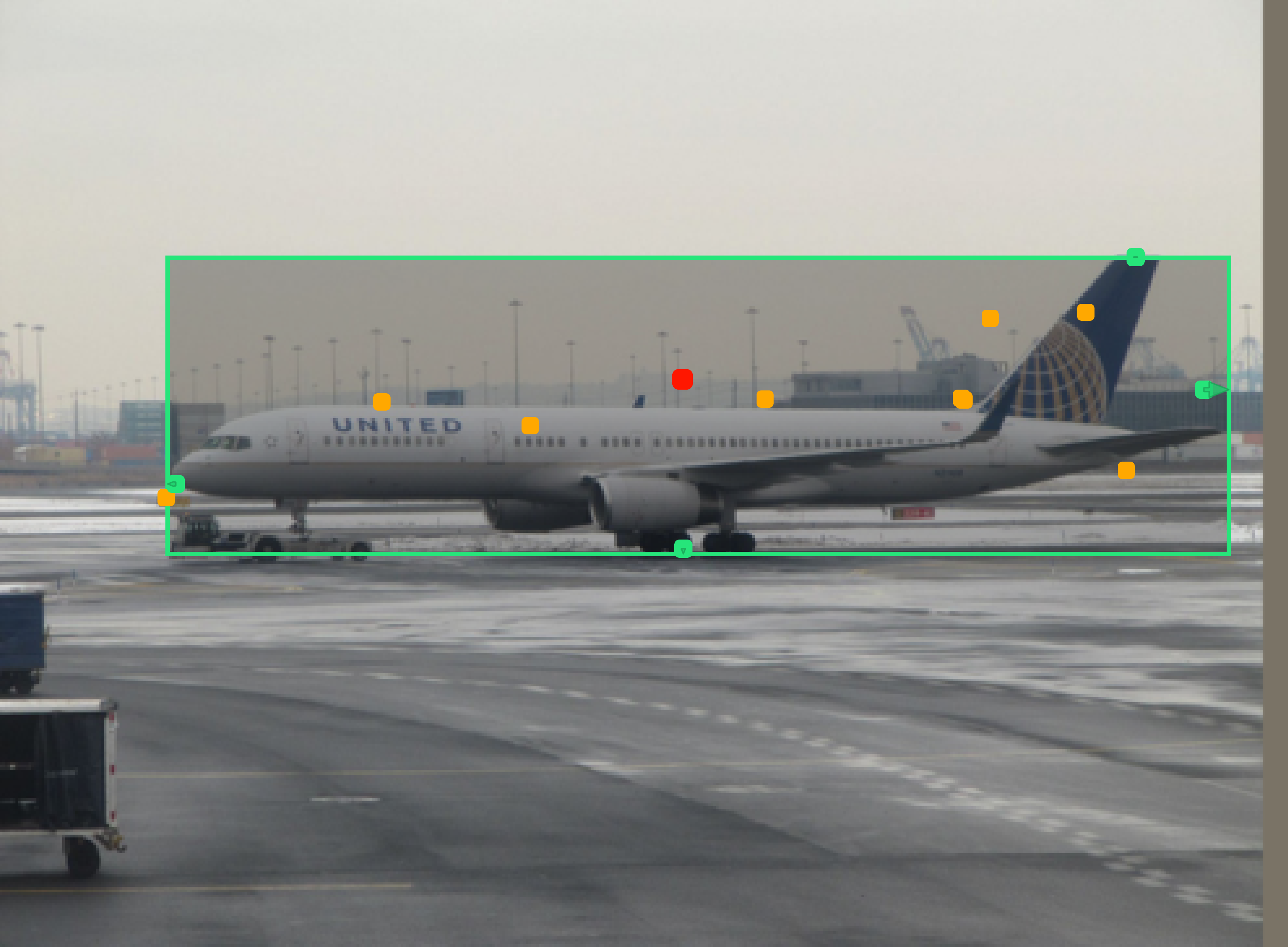}
	\end{minipage}%
	\\
	\begin{minipage}[t]{0.998\textwidth}
		\centering
		\includegraphics[height=0.222\textwidth, trim=0 0 20 0,clip]{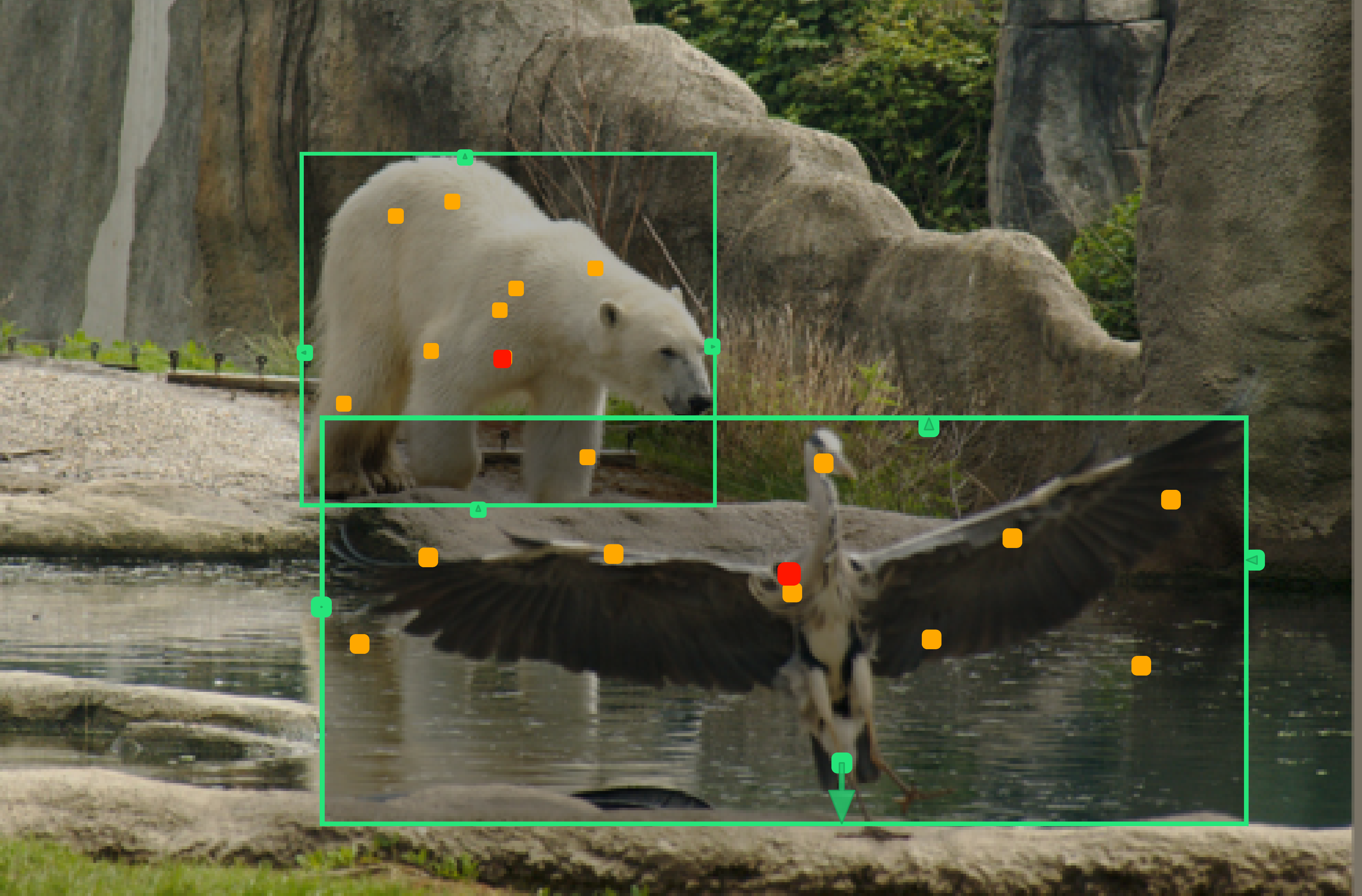}
		\includegraphics[height=0.222\textwidth, trim=0 0 20 0,clip]{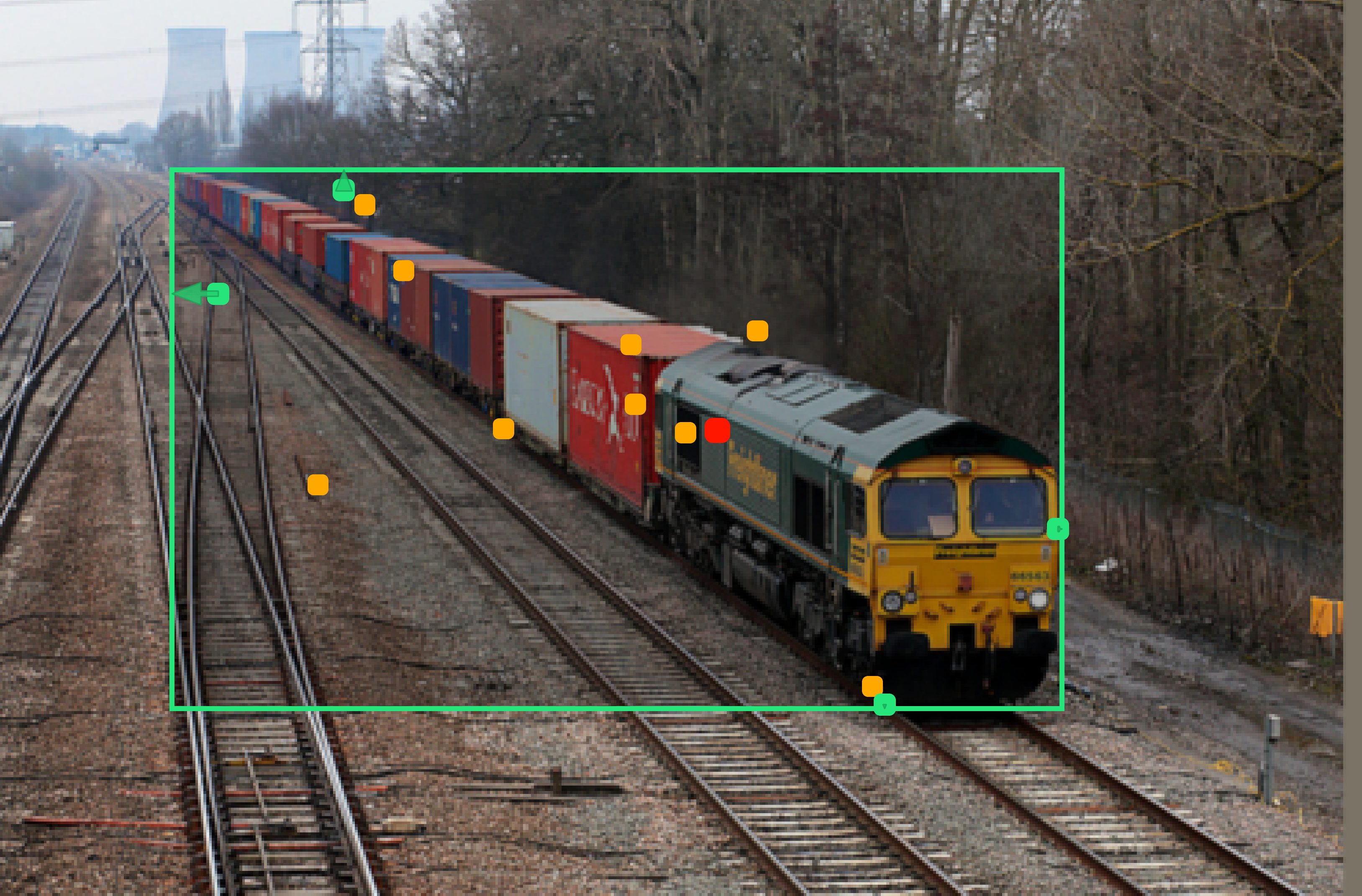}
		\includegraphics[height=0.222\textwidth, trim=0 20 0 0,clip]{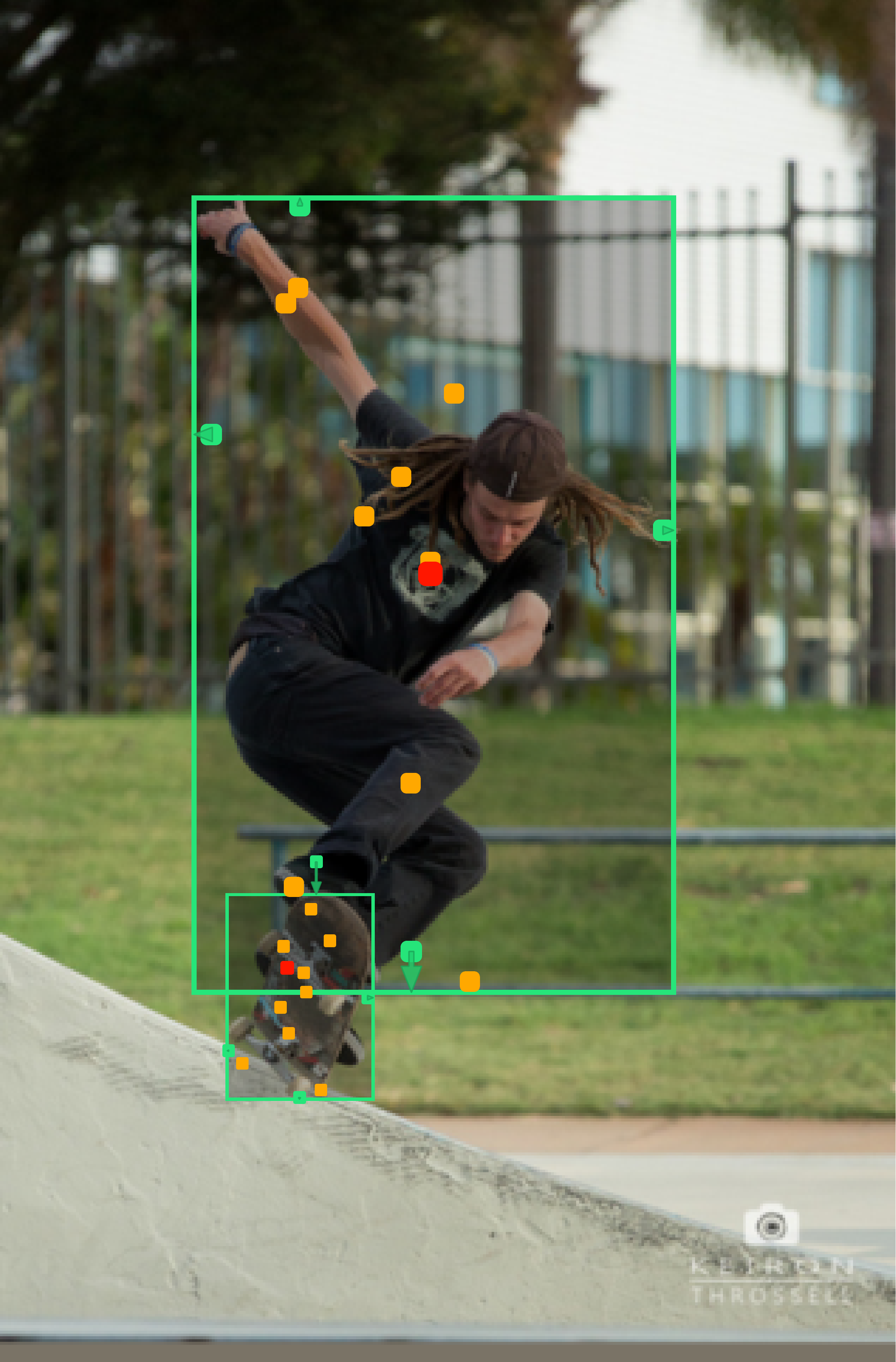}
		\includegraphics[height=0.222\textwidth, trim=0 20 0 0,clip]{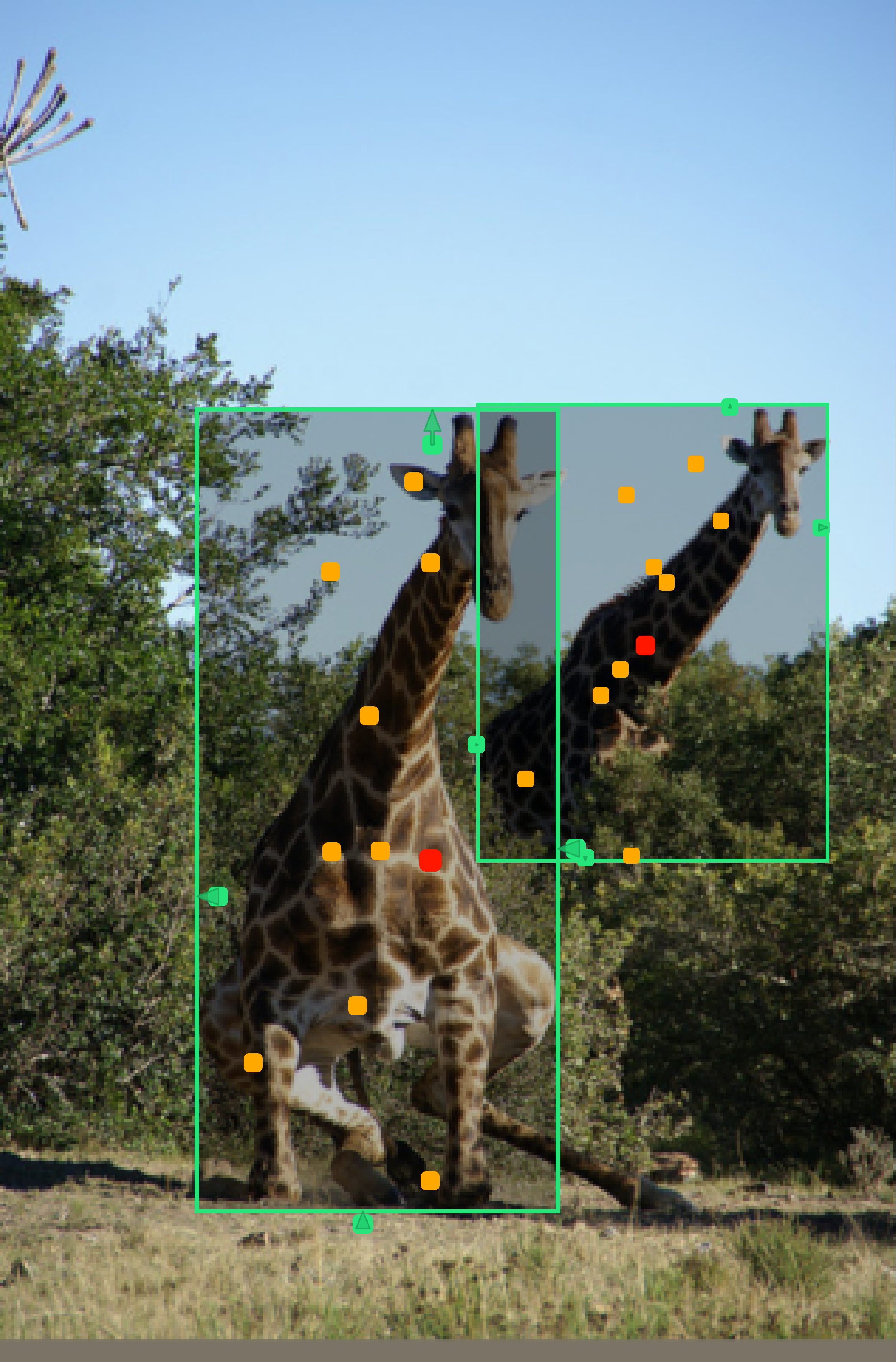}
	\end{minipage}%
	\caption{Visualization of some detection results on the MS COCO \texttt{val2017} set. 
	The \revised{detected} object bounding boxes are illustrated in green, and the predicted dynamic boundary points and semantic points are plotted in green and orange, respectively. 
	\revised{We also mark the source grids that generate these dynamic points in red, and use green arrows to indicate the regression offsets collected from the dynamic boundary points.} 
	The dynamic boundary points (green) are \revised{generally} located near the object edges, where the bounding box boundaries can be accurately inferred. %\sout{we can obtain accurate}\sout{localization} \sout{results}. 
	The dynamic semantic points (orange) are mainly distributed over different parts of the object, which is advantageous to object classification. %\sout{help}\sout{to identify\yx{ing} the object categories}
	}
	\label{fig_vis_dets}
\end{figure*}

In Fig.~\ref{fig_vis_dets}, we demonstrate some detection results on the MS COCO \texttt{val2017} split~\cite{lin2014microsoft}. %\sout{, as well as the dynamic boundary points (in green) and semantic points (in yellow) used to produce these detection results}. 
\revised{Specifically, for each image, we apply our PDNet model to produce a set of detected objects. Then, we collect the dynamic boundary points and semantic points that produce these detection results, and map them all to the original image for visualization. The detected object bounding boxes are illustrated in green, and the dynamic boundary points and semantic points are plotted in green and orange, respectively. We also mark the source grids that generate these dynamic points in red, and use green arrows to indicate the regression offsets collected from the dynamic boundary points.} 
The visualization results manifest that the predicted dynamic boundary points are located %\sout{close to} \sout{objects' horizontal or vertical edges}\sout{helping to} 
near the object edges where the bounding box boundaries can be better inferred, %where the object boundary can be better inferred, %\sout{to better obtain the precise boundary localization \sout{results}}, 
and the dynamic semantic points are more likely to disperse over different object parts %\sout{mostly} \sout{distributed}areas of the objects  adapting to various shapes of object
to collect more reasonable classification predictions. %\sout{effective} crucial, which helps to better identify the objects. 
For example, in the first image of Fig.~\ref{fig_vis_dets}, the leftmost border of the cat is accurately localized with a dynamic boundary point on its tail, %a boundary point is near the cat's tail to accurately localize the leftmost boundary,
while the semantic points tend to scatter over the cat's body to comprehensively classify the cat. % \sout{are} \sout{mainly scattered}\sout{on} \sout{classification}

\subsection{Efficiency}

\begin{table}[t]
    \renewcommand{\arraystretch}{1.148}
    \caption{Comparison of the inference time.} % of our method and other detectors.}
	%\centering
	\begin{center}
	\resizebox{0.918\columnwidth}{!}{
	\begin{tabular}{l|c|c|c|c}
		\hline
        Method & Backbone & $\mathrm{AP}$ & FPS & Test time (ms) \\ %& Batch size\\
		\hline
		FCOS~\cite{tian2019fcos}                & ResNet-50         & 38.7  & 16.6  & 60.1 \\
		ATSS~\cite{zhang2020bridging}           & ResNet-50         & 39.3  & 16.6  & 60.1 \\
		\hline
		RepPoints v2~\cite{chen2020reppoints}   & ResNet-50         & 41.0  & 9.7  & 103.2 \\
		BorderDet~\cite{qiu2020borderdet}       & ResNet-50         & 41.4  & 11.7  & 85.2 \\
		\hline
		%\textit{\footnotesize Our test:} & & & \\
		PDNet & ResNet-50 & 41.8  & 16.2 & 61.8 \\
		\hline
	\end{tabular}
	}
	\end{center}
	\label{table_efficiency}
	\vspace{-0.2cm}
\end{table}

We evaluate the inference time of our proposed PDNet and other recent dense detection methods for efficiency comparison. All the experiments are conducted using a single NVIDIA 1080Ti GPU. As shown in Table~\ref{table_efficiency}, our detector runs  %\sout{with a similar inference time to the}
almost as fast as the one-stage detection methods FCOS~\cite{tian2019fcos} and ATSS~\cite{zhang2020bridging}, since our prediction collection process is lightweight and incurs negligible overhead. 
Compared with the recent RepPoints v2~\cite{chen2020reppoints} and BorderDet~\cite{qiu2020borderdet} with more complicated detection heads, 
%The recently proposed RepPoints v2~\cite{chen2020reppoints} and BorderDet~\cite{qiu2020borderdet} achieve leading detection performance with more complicated detection pipelines. Compared with them, 
our method achieves a higher AP while being significantly faster, which demonstrates the advantages of our method in both efficiency and precision. %\sout{accuracy}  

\section{Conclusion}
In this work, we %\sout{have proposed}
propose an accurate and efficient object detector PDNet that infers different targets (\ie, the object category and boundary locations) at their corresponding appropriate positions. 
Specifically, based on the dense prediction approach, we propose the PDNet with a prediction decoupling mechanism to flexibly collect different target predictions from different locations and aggregate them for the final detection results. %\sout{the predictions} \sout{for different targets} from the dense regression and classification prediction maps. 
Moreover, we devise two sets of dynamic points, \ie, dynamic boundary points and semantic points, and incorporate a two-step generation strategy to facilitate the learning of suitable inference positions for localization and classification.
Extensive experiments on the MS COCO benchmark demonstrate the state-of-the-art performance and efficiency %\sout{advantage} 
of our method.

\section*{Acknowledgment}

This work was supported by the National Key Research and Development Program of China (Grant No. 2020AAA0106800), Beijing Natural Science Foundation (Grant No. JQ21017, 4224091), the Natural Science Foundation of China (Grant No. 61972397, 62036011, 62192782, 61721004, 61906192), the Key Research Program of Frontier Sciences, CAS, Grant No. QYZDJ-SSW-JSC040, and the China Postdoctoral Science Foundation (Grant No. 2021M693402).


% \clearpage
\ifCLASSOPTIONcaptionsoff
  \newpage
\fi



\bibliographystyle{IEEEtran}
\bibliography{IEEEabrv,ref}

\begin{thebibliography}{10}
\providecommand{\url}[1]{#1}
\csname url@samestyle\endcsname
\providecommand{\newblock}{\relax}
\providecommand{\bibinfo}[2]{#2}
\providecommand{\BIBentrySTDinterwordspacing}{\spaceskip=0pt\relax}
\providecommand{\BIBentryALTinterwordstretchfactor}{4}
\providecommand{\BIBentryALTinterwordspacing}{\spaceskip=\fontdimen2\font plus
\BIBentryALTinterwordstretchfactor\fontdimen3\font minus
  \fontdimen4\font\relax}
\providecommand{\BIBforeignlanguage}[2]{{%
\expandafter\ifx\csname l@#1\endcsname\relax
\typeout{** WARNING: IEEEtran.bst: No hyphenation pattern has been}%
\typeout{** loaded for the language `#1'. Using the pattern for}%
\typeout{** the default language instead.}%
\else
\language=\csname l@#1\endcsname
\fi
#2}}
\providecommand{\BIBdecl}{\relax}
\BIBdecl

\bibitem{liu2016ssd}
W.~Liu, D.~Anguelov, D.~Erhan, C.~Szegedy, S.~Reed, C.-Y. Fu, and A.~C. Berg,
  ``Ssd: Single shot multibox detector,'' in \emph{European conference on
  computer vision}.\hskip 1em plus 0.5em minus 0.4em\relax Springer, 2016, pp.
  21--37.

\bibitem{lin2017focal}
T.-Y. Lin, P.~Goyal, R.~Girshick, K.~He, and P.~Doll{\'a}r, ``Focal loss for
  dense object detection,'' in \emph{Proceedings of the IEEE international
  conference on computer vision}, 2017, pp. 2980--2988.

\bibitem{ren2015faster}
S.~Ren, K.~He, R.~Girshick, and J.~Sun, ``Faster r-cnn: Towards real-time
  object detection with region proposal networks,'' in \emph{Advances in neural
  information processing systems}, 2015, pp. 91--99.

\bibitem{dai2016r}
J.~Dai, Y.~Li, K.~He, and J.~Sun, ``R-fcn: Object detection via region-based
  fully convolutional networks,'' in \emph{Advances in neural information
  processing systems}, 2016, pp. 379--387.

\bibitem{tian2019fcos}
Z.~Tian, C.~Shen, H.~Chen, and T.~He, ``Fcos: Fully convolutional one-stage
  object detection,'' in \emph{Proceedings of the IEEE international conference
  on computer vision}, 2019, pp. 9627--9636.

\bibitem{kong2020foveabox}
T.~Kong, F.~Sun, H.~Liu, Y.~Jiang, L.~Li, and J.~Shi, ``Foveabox: Beyound
  anchor-based object detection,'' \emph{IEEE Transactions on Image
  Processing}, vol.~29, pp. 7389--7398, 2020.

\bibitem{zhu2019feature}
C.~Zhu, Y.~He, and M.~Savvides, ``Feature selective anchor-free module for
  single-shot object detection,'' in \emph{Proceedings of the IEEE/CVF
  Conference on Computer Vision and Pattern Recognition}, 2019, pp. 840--849.

\bibitem{zhu2020soft}
C.~Zhu, F.~Chen, Z.~Shen, and M.~Savvides, ``Soft anchor-point object
  detection,'' in \emph{European conference on computer vision}.\hskip 1em plus
  0.5em minus 0.4em\relax Springer, 2020, pp. 91--107.

\bibitem{girshick2014rich}
R.~Girshick, J.~Donahue, T.~Darrell, and J.~Malik, ``Rich feature hierarchies
  for accurate object detection and semantic segmentation,'' in
  \emph{Proceedings of the IEEE conference on computer vision and pattern
  recognition}, 2014, pp. 580--587.

\bibitem{girshick2015fast}
R.~Girshick, ``Fast r-cnn,'' in \emph{Proceedings of the IEEE international
  conference on computer vision}, 2015, pp. 1440--1448.

\bibitem{lin2017feature}
T.-Y. Lin, P.~Doll{\'a}r, R.~Girshick, K.~He, B.~Hariharan, and S.~Belongie,
  ``Feature pyramid networks for object detection,'' in \emph{Proceedings of
  the IEEE conference on computer vision and pattern recognition}, 2017, pp.
  2117--2125.

\bibitem{li2019scale}
Y.~Li, Y.~Chen, N.~Wang, and Z.~Zhang, ``Scale-aware trident networks for
  object detection,'' in \emph{Proceedings of the IEEE/CVF International
  Conference on Computer Vision}, 2019, pp. 6054--6063.

\bibitem{cai2018cascade}
Z.~Cai and N.~Vasconcelos, ``Cascade r-cnn: Delving into high quality object
  detection,'' in \emph{Proceedings of the IEEE conference on computer vision
  and pattern recognition}, 2018, pp. 6154--6162.

\bibitem{he2017mask}
K.~He, G.~Gkioxari, P.~Doll{\'a}r, and R.~Girshick, ``Mask r-cnn,'' in
  \emph{Proceedings of the IEEE international conference on computer vision},
  2017, pp. 2961--2969.

\bibitem{dai2017deformable}
J.~Dai, H.~Qi, Y.~Xiong, Y.~Li, G.~Zhang, H.~Hu, and Y.~Wei, ``Deformable
  convolutional networks,'' in \emph{Proceedings of the IEEE international
  conference on computer vision}, 2017, pp. 764--773.

\bibitem{zhu2019deformable}
X.~Zhu, H.~Hu, S.~Lin, and J.~Dai, ``Deformable convnets v2: More deformable,
  better results,'' in \emph{Proceedings of the IEEE Conference on Computer
  Vision and Pattern Recognition}, 2019, pp. 9308--9316.

\bibitem{wang2019region}
J.~Wang, K.~Chen, S.~Yang, C.~C. Loy, and D.~Lin, ``Region proposal by guided
  anchoring,'' in \emph{Proceedings of the IEEE Conference on Computer Vision
  and Pattern Recognition}, 2019, pp. 2965--2974.

\bibitem{vu2019cascade}
T.~Vu, H.~Jang, T.~X. Pham, and C.~Yoo, ``Cascade rpn: Delving into
  high-quality region proposal network with adaptive convolution,'' in
  \emph{Advances in Neural Information Processing Systems}, 2019, pp.
  1432--1442.

\bibitem{shrivastava2016training}
A.~Shrivastava, A.~Gupta, and R.~Girshick, ``Training region-based object
  detectors with online hard example mining,'' in \emph{Proceedings of the IEEE
  conference on computer vision and pattern recognition}, 2016, pp. 761--769.

\bibitem{singh2018analysis}
B.~Singh and L.~S. Davis, ``An analysis of scale invariance in object detection
  snip,'' in \emph{Proceedings of the IEEE conference on computer vision and
  pattern recognition}, 2018, pp. 3578--3587.

\bibitem{jiang2018acquisition}
B.~Jiang, R.~Luo, J.~Mao, T.~Xiao, and Y.~Jiang, ``Acquisition of localization
  confidence for accurate object detection,'' in \emph{Proceedings of the
  European Conference on Computer Vision (ECCV)}, 2018, pp. 784--799.

\bibitem{pang2019libra}
J.~Pang, K.~Chen, J.~Shi, H.~Feng, W.~Ouyang, and D.~Lin, ``Libra r-cnn:
  Towards balanced learning for object detection,'' in \emph{Proceedings of the
  IEEE/CVF Conference on Computer Vision and Pattern Recognition}, 2019, pp.
  821--830.

\bibitem{zhang2020dynamic}
H.~Zhang, H.~Chang, B.~Ma, N.~Wang, and X.~Chen, ``Dynamic r-cnn: Towards high
  quality object detection via dynamic training,'' in \emph{European Conference
  on Computer Vision}.\hskip 1em plus 0.5em minus 0.4em\relax Springer, 2020,
  pp. 260--275.

\bibitem{he2019bounding}
Y.~He, C.~Zhu, J.~Wang, M.~Savvides, and X.~Zhang, ``Bounding box regression
  with uncertainty for accurate object detection,'' in \emph{Proceedings of the
  IEEE/CVF Conference on Computer Vision and Pattern Recognition}, 2019, pp.
  2888--2897.

\bibitem{fu2017dssd}
C.-Y. Fu, W.~Liu, A.~Ranga, A.~Tyagi, and A.~C. Berg, ``Dssd: Deconvolutional
  single shot detector,'' \emph{arXiv preprint arXiv:1701.06659}, 2017.

\bibitem{zhang2018single}
S.~Zhang, L.~Wen, X.~Bian, Z.~Lei, and S.~Z. Li, ``Single-shot refinement
  neural network for object detection,'' in \emph{Proceedings of the IEEE
  conference on computer vision and pattern recognition}, 2018, pp. 4203--4212.

\bibitem{zhang2019freeanchor}
X.~Zhang, F.~Wan, C.~Liu, R.~Ji, and Q.~Ye, ``Freeanchor: Learning to match
  anchors for visual object detection,'' in \emph{Advances in Neural
  Information Processing Systems}, 2019, pp. 147--155.

\bibitem{ke2020multiple}
W.~Ke, T.~Zhang, Z.~Huang, Q.~Ye, J.~Liu, and D.~Huang, ``Multiple anchor
  learning for visual object detection,'' in \emph{Proceedings of the IEEE/CVF
  Conference on Computer Vision and Pattern Recognition}, 2020, pp.
  10\,206--10\,215.

\bibitem{zhang2020bridging}
S.~Zhang, C.~Chi, Y.~Yao, Z.~Lei, and S.~Z. Li, ``Bridging the gap between
  anchor-based and anchor-free detection via adaptive training sample
  selection,'' in \emph{Proceedings of the IEEE/CVF Conference on Computer
  Vision and Pattern Recognition}, 2020, pp. 9759--9768.

\bibitem{li2020generalized}
X.~Li, W.~Wang, L.~Wu, S.~Chen, X.~Hu, J.~Li, J.~Tang, and J.~Yang,
  ``Generalized focal loss: Learning qualified and distributed bounding boxes
  for dense object detection,'' in \emph{NeurIPS}, 2020.

\bibitem{redmon2016you}
J.~Redmon, S.~Divvala, R.~Girshick, and A.~Farhadi, ``You only look once:
  Unified, real-time object detection,'' in \emph{Proceedings of the IEEE
  conference on computer vision and pattern recognition}, 2016, pp. 779--788.

\bibitem{huang2015densebox}
L.~Huang, Y.~Yang, Y.~Deng, and Y.~Yu, ``Densebox: Unifying landmark
  localization with end to end object detection,'' \emph{arXiv preprint
  arXiv:1509.04874}, 2015.

\bibitem{wu2020rethinking}
Y.~Wu, Y.~Chen, L.~Yuan, Z.~Liu, L.~Wang, H.~Li, and Y.~Fu, ``Rethinking
  classification and localization for object detection,'' in \emph{Proceedings
  of the IEEE/CVF conference on computer vision and pattern recognition}, 2020,
  pp. 10\,186--10\,195.

\bibitem{song2020revisiting}
G.~Song, Y.~Liu, and X.~Wang, ``Revisiting the sibling head in object
  detector,'' in \emph{Proceedings of the IEEE/CVF Conference on Computer
  Vision and Pattern Recognition}, 2020, pp. 11\,563--11\,572.

\bibitem{lu2019grid}
X.~Lu, B.~Li, Y.~Yue, Q.~Li, and J.~Yan, ``Grid r-cnn,'' in \emph{Proceedings
  of the IEEE/CVF Conference on Computer Vision and Pattern Recognition}, 2019,
  pp. 7363--7372.

\bibitem{wang2020side}
J.~Wang, W.~Zhang, Y.~Cao, K.~Chen, J.~Pang, T.~Gong, J.~Shi, C.~C. Loy, and
  D.~Lin, ``Side-aware boundary localization for more precise object
  detection,'' in \emph{European Conference on Computer Vision}.\hskip 1em plus
  0.5em minus 0.4em\relax Springer, 2020, pp. 403--419.

\bibitem{yang2019reppoints}
Z.~Yang, S.~Liu, H.~Hu, L.~Wang, and S.~Lin, ``Reppoints: Point set
  representation for object detection,'' in \emph{Proceedings of the IEEE
  International Conference on Computer Vision}, 2019, pp. 9657--9666.

\bibitem{chen2020reppoints}
Y.~Chen, Z.~Zhang, Y.~Cao, L.~Wang, S.~Lin, and H.~Hu, ``Reppoints v2:
  Verification meets regression for object detection,'' \emph{Advances in
  Neural Information Processing Systems}, vol.~33, 2020.

\bibitem{chen2019revisiting}
Y.~Chen, C.~Han, N.~Wang, and Z.~Zhang, ``Revisiting feature alignment for
  one-stage object detection,'' \emph{arXiv preprint arXiv:1908.01570}, 2019.

\bibitem{qiu2020borderdet}
H.~Qiu, Y.~Ma, Z.~Li, S.~Liu, and J.~Sun, ``Borderdet: Border feature for dense
  object detection,'' in \emph{European Conference on Computer Vision}.\hskip
  1em plus 0.5em minus 0.4em\relax Springer, 2020, pp. 549--564.

\bibitem{law2018cornernet}
H.~Law and J.~Deng, ``Cornernet: Detecting objects as paired keypoints,'' in
  \emph{Proceedings of the European conference on computer vision (ECCV)},
  2018, pp. 734--750.

\bibitem{duan2019centernet}
K.~Duan, S.~Bai, L.~Xie, H.~Qi, Q.~Huang, and Q.~Tian, ``Centernet: Keypoint
  triplets for object detection,'' in \emph{Proceedings of the IEEE/CVF
  International Conference on Computer Vision}, 2019, pp. 6569--6578.

\bibitem{zhou2019objects}
X.~Zhou, D.~Wang, and P.~Kr{\"a}henb{\"u}hl, ``Objects as points,'' \emph{arXiv
  preprint arXiv:1904.07850}, 2019.

\bibitem{zhou2019bottom}
X.~Zhou, J.~Zhuo, and P.~Krahenbuhl, ``Bottom-up object detection by grouping
  extreme and center points,'' in \emph{Proceedings of the IEEE/CVF Conference
  on Computer Vision and Pattern Recognition}, 2019, pp. 850--859.

\bibitem{lin2014microsoft}
T.-Y. Lin, M.~Maire, S.~Belongie, J.~Hays, P.~Perona, D.~Ramanan,
  P.~Doll{\'a}r, and C.~L. Zitnick, ``Microsoft coco: Common objects in
  context,'' in \emph{European conference on computer vision}.\hskip 1em plus
  0.5em minus 0.4em\relax Springer, 2014, pp. 740--755.

\bibitem{he2016deep}
K.~He, X.~Zhang, S.~Ren, and J.~Sun, ``Deep residual learning for image
  recognition,'' in \emph{Proceedings of the IEEE conference on computer vision
  and pattern recognition}, 2016, pp. 770--778.

\bibitem{rezatofighi2019generalized}
H.~Rezatofighi, N.~Tsoi, J.~Gwak, A.~Sadeghian, I.~Reid, and S.~Savarese,
  ``Generalized intersection over union: A metric and a loss for bounding box
  regression,'' in \emph{Proceedings of the IEEE Conference on Computer Vision
  and Pattern Recognition}, 2019, pp. 658--666.

\bibitem{deng2009imagenet}
J.~Deng, W.~Dong, R.~Socher, L.-J. Li, K.~Li, and L.~Fei-Fei, ``Imagenet: A
  large-scale hierarchical image database,'' in \emph{2009 IEEE conference on
  computer vision and pattern recognition}.\hskip 1em plus 0.5em minus
  0.4em\relax Ieee, 2009, pp. 248--255.

\end{thebibliography}
\end{document}